\documentclass{article}

\usepackage{microtype}
\usepackage{graphicx}
\usepackage{subfigure}
\usepackage{booktabs} % for professional tables

\usepackage{hyperref}

\usepackage[accepted]{icml2023}

\usepackage{amsmath}
\usepackage{amssymb}
\usepackage{mathtools}
\usepackage{amsthm}

\usepackage[capitalize,noabbrev]{cleveref}

\usepackage{url}            % simple URL typesetting
\usepackage{booktabs}       % professional-quality tables
\usepackage{amsfonts}       % blackboard math symbols
\usepackage{nicefrac}       % compact symbols for 1/2, etc.
\usepackage{microtype}      % microtypography
\usepackage{xcolor}         % colors
\usepackage{amsmath}
\usepackage{amssymb}
\usepackage{mathtools}
\usepackage{amsthm}
\usepackage{bm}
\usepackage{graphicx}
\usepackage{subfigure}
\RequirePackage{algorithm}
\RequirePackage{algorithmic}
\usepackage{paralist}
\usepackage{comment}
\usepackage{macros}
\usepackage{adjustbox}
\usepackage{wrapfig}

\usepackage{enumitem}
\usepackage{soul}

\usepackage{tikz}
\usepackage{tikzit}
\tikzstyle{basic}=[fill=white, draw=black, shape=circle]
\tikzstyle{square}=[fill=white, draw=black, shape=rectangle]
\tikzstyle{big dashed}=[fill=white, draw=black, shape=circle, minimum width=1cm, dashed]
\tikzstyle{vertical ellipse dashed}=[fill=none, draw=blue, minimum width=0.75cm, minimum height=3cm, ellipse, dashed, tikzit shape=rectangle, tikzit draw=blue, tikzit fill=white]
\tikzstyle{small vertical ellipse dashed}=[fill=none, draw=blue, shape=circle, tikzit fill=white, tikzit draw=blue, dashed, minimum width=0.75cm, minimum height=1.5cm, tikzit shape=rectangle, ellipse]
\tikzstyle{tiny vertical ellipse dashed}=[fill=none, draw=blue, shape=circle, tikzit fill=white, ellipse, dashed, minimum width=0.75cm, minimum height=1cm, tikzit shape=rectangle]
\tikzstyle{red}=[fill=red, draw=black, shape=circle]
\tikzstyle{green}=[fill={rgb,255: red,0; green,128; blue,128}, draw=black, shape=circle]
\tikzstyle{blue}=[fill=blue, draw=black, shape=circle]
\tikzstyle{huge dashed}=[fill=white, draw=black, shape=circle, dashed, minimum width=2cm]
\tikzstyle{medium}=[fill=white, draw=black, shape=circle, minimum width=1cm]
\tikzstyle{pale green}=[fill={rgb,255: red,173; green,231; blue,0}, draw=black, shape=circle, minimum width=1cm]
\tikzstyle{horizontal ellipse dashed}=[fill=white, draw=black, tikzit draw=magenta, tikzit shape=rectangle, minimum width=3cm, minimum height=0.75cm, ellipse, dashed]
\tikzstyle{minsize}=[fill=white, draw=black, shape=circle, minimum width=0.75cm]
\tikzstyle{horizontal ellipse green}=[fill={rgb,255: red,191; green,255; blue,0}, draw=black, tikzit draw={rgb,255: red,191; green,255; blue,0}, tikzit shape=rectangle, minimum width=3cm, minimum height=0.75cm, ellipse, dashed]
\tikzstyle{horizontal ellipse blue}=[fill={rgb,255: red,107; green,203; blue,255}, draw=black, tikzit draw=blue, tikzit shape=rectangle, minimum width=3cm, minimum height=0.75cm, ellipse, dashed]
\tikzstyle{smallblack}=[fill=black, draw=black, shape=circle, inner sep=0 pt, minimum size=3 pt]
\tikzstyle{smallSquare}=[fill=white, draw=black, shape=rectangle, inner sep=0 pt, minimum size=6 pt]
\tikzstyle{smallCircle}=[fill=white, draw=black, shape=circle, inner sep=0 pt, minimum size=6 pt]
\tikzstyle{big vertical ellipse dashed}=[fill=none, draw=blue, shape=circle, tikzit shape=rectangle, ellipse, dashed, minimum width=0.95cm, minimum height=3.7cm]
\tikzstyle{smallred}=[fill=red, draw=red, shape=circle, inner sep=0 pt, minimum size=3 pt]
\tikzstyle{redfilled}=[fill={red!20}, draw=red, shape=circle, opacity=0.5]
\tikzstyle{bluefilled}=[fill={blue!20}, draw=blue, shape=circle, opacity=0.5]
\tikzstyle{greenfilled}=[fill={rgb,255: red,149; green,255; blue,179}, draw={rgb,255: red,0; green,107; blue,61}, shape=circle, opacity=0.5]
\tikzstyle{orangefilled}=[fill={orange!20}, draw=orange, shape=circle, opacity=0.5]
\tikzstyle{new style 0}=[fill={rgb,255: red,191; green,191; blue,191}, draw=black, shape=circle]
\tikzstyle{small pink}=[fill=white, draw={rgb,255: red,18; green,162; blue,206}, shape=circle, inner sep=0 pt, minimum size=6 pt]
\tikzstyle{smallred}=[fill=white, draw=red, shape=circle, inner sep=0 pt, minimum size=6 pt]

\tikzstyle{directed}=[->, -latex, draw=black]
\tikzstyle{undirected}=[-, draw={rgb,255: red,128; green,128; blue,128}]
\tikzstyle{directed red}=[draw=red, ->, -latex]
\tikzstyle{directed green}=[draw={rgb,255: red,0; green,128; blue,128}, ->, line width=1pt]
\tikzstyle{directed blue}=[draw=blue, ->, line width=1pt]
\tikzstyle{directed purple}=[draw={rgb,255: red,128; green,0; blue,128}, ->, line width=1pt]
\tikzstyle{undirected red}=[-, draw=red]
\tikzstyle{undirected green}=[-, draw={rgb,255: red,0; green,107; blue,61}, line width=1pt]
\tikzstyle{undirected blue}=[-, draw=blue, line width=1pt]
\tikzstyle{undirected purple}=[-, draw={rgb,255: red,128; green,0; blue,128}, line width=1pt]
\tikzstyle{undirected dashed}=[-, line width=1pt, dashed, draw=black]
\tikzstyle{orange dashed}=[-, draw={rgb,255: red,255; green,128; blue,0}, dashed, line width=1.5pt]
\tikzstyle{directed dash}=[->, dashed]
\tikzstyle{blue dashed}=[-, draw=blue, dashed, line width=1pt]
\tikzstyle{green dashed}=[-, draw={rgb,255: red,0; green,162; blue,0}, dashed, line width=1pt]
\tikzstyle{blue filled}=[-, fill={blue!20}, draw=blue, line width=1pt, opacity=0.5, tikzit fill=white]
\tikzstyle{red filled}=[-, fill={red!20}, line width=1pt, draw=red, opacity=0.5, tikzit fill=white]
\tikzstyle{green filled}=[-, line width=1pt, draw={rgb,255: red,0; green,107; blue,61}, opacity=0.5, tikzit fill=white, fill={rgb,255: red,149; green,255; blue,179}]
\tikzstyle{orange filled}=[-, fill={orange!20}, draw=orange, line width=1pt, opacity=0.5, tikzit fill=white]
\tikzstyle{undirected dashed}=[-, draw={rgb,255: red,128; green,128; blue,128}, dashed, line width=1pt]
\tikzstyle{directed}=[->, -latex, fill=none, draw={rgb,255: red,128; green,128; blue,128}]
\tikzstyle{reddashed}=[-, draw=red, dashed]
\tikzstyle{blackdashed}=[-, dashed, draw=black]
\tikzstyle{black}=[-, ->, -latex, draw=black]
\tikzstyle{pinksolid}=[-, draw={rgb,255: red,18; green,162; blue,206}]
\tikzstyle{pinkdirected}=[-, ->, -latex, draw={rgb,255: red,18; green,162; blue,206}]
\tikzstyle{pink dashed}=[-, dashed, draw={rgb,255: red,18; green,162; blue,206}]

\tikzstyle{swr}=[fill=white, draw=black, shape=rectangle]
\tikzstyle{swc}=[fill=white, draw=black, shape=circle, minimum size=0.5 cm, inner sep=0.02cm]
\tikzstyle{mwc}=[fill=white, draw=black, shape=circle, minimum size=0.75cm, inner sep=0.02cm]
\tikzstyle{oval}=[fill=white, draw=black, shape=circle, minimum height=0.5cm, minimum width=1 cm, ellipse, inner sep=0.02cm]
\tikzstyle{mbc}=[fill={rgb,255: red,191; green,191; blue,191}, draw=black, shape=circle, minimum size=0.75cm, inner sep=0.02cm]

\tikzstyle{sb}=[-]
\tikzstyle{new edge style 0}=[-, draw=red]

\theoremstyle{plain}

\theoremstyle{definition}

\theoremstyle{remark}

\usepackage[textsize=small]{todonotes}

\icmltitlerunning{Parallel Neurosymbolic Integration with Concordia}

\begin{document}

\twocolumn[
\icmltitle{Parallel Neurosymbolic Integration with Concordia}



\icmlsetsymbol{equal}{*}

\begin{icmlauthorlist}
\icmlauthor{Jonathan Feldstein}{equal,yyy,comp1}
\icmlauthor{Modestas Jur\v{c}ius}{equal,yyy,comp2}
\icmlauthor{Efthymia Tsamoura}{comp3}
\end{icmlauthorlist}

\icmlaffiliation{yyy}{University of Edinburgh, Edinburgh, United Kingdom}
\icmlaffiliation{comp1}{BENNU.AI, Edinburgh, United Kingdom}
\icmlaffiliation{comp2}{Mintis AI, Kaunas, Lithuania}
\icmlaffiliation{comp3}{Samsung AI, Cambridge, United Kingdom}

\icmlcorrespondingauthor{Jonathan Feldstein}{jonathan.feldstein@bennu.ai}

\icmlkeywords{Machine Learning, ICML, Neuro-Symbolic AI}

\vskip 0.3in
]



\printAffiliationsAndNotice{\icmlEqualContribution} % otherwise use the standard text.

\begin{abstract}
Parallel neurosymbolic architectures have been applied 
effectively in NLP by distilling knowledge from a logic theory into a deep model.
However, prior art faces several limitations including supporting restricted forms of logic theories and relying on the assumption of independence between the logic and the deep network. We present Concordia, a framework overcoming the limitations of prior art. Concordia is agnostic both to the deep network and the logic theory offering support for a wide range of probabilistic theories. Our framework can support supervised training of both components and unsupervised training of the neural component. Concordia has been successfully applied to tasks beyond NLP and data classification, improving the accuracy of state-of-the-art 
on collective activity detection, entity linking and recommendation tasks.
\end{abstract}

\section{Introduction} \label{section:introduction}

\textbf{Motivation.} 
To overcome the limitations of deep networks, such as dependence on significant amount of labelled training data, researchers proposed to integrate logical theories, a computational paradigm known as \emph{neurosymbolic AI} \cite{DBLP:books/daglib/0007534}. 
One way to integrate the components is in a staged or \emph{stratified} fashion.
Stratified neurosymbolic frameworks find applications in problems admitting well-separable symbolic and subsymbolic tasks – to name a toy example, performing mathematical operations using symbolic models over MNIST digits identified by a neural model. One of the first stratified architectures was DeepProbLog \cite{deepproblog}. More followed: NeurASP \cite{neurasp}, ABL \cite{abl}, RNMs \cite{DBLP:conf/ecai/MarraDGGM20} and NeuroLog \cite{tsamoura2020neuralsymbolic}.

An alternative to stratified is \textit{parallel} integration. 
In contrast to stratified frameworks, parallel integration applies in settings in which the same task can be solved both symbolically and sub-symbolically and the aim is to increase the accuracy of the end task by distilling knowledge from the logic component into the neural one and vice versa. 
Two parallel neurosymbolic frameworks have been proposed recently: Teacher-Student (T-S) by Hu et al. \cite{hu-etal-2016-harnessing,hu-etal-2016-deep} and Deep Probabilistic Logic (DPL) by Wang and Poon \cite{DBLP:conf/emnlp/WangP18}. T-S is based on posterior regularization \cite{posterior} to build a teacher network which is later used to train the neural module. DPL defines a joint distribution after making the assumption of independence between the logical theory and the deep network and uses this distribution to regularize the deep model.   

\textbf{Problem.} Stratified approaches focus on separating the pattern recognition and the reasoning to allow for more complex reasoning and querying. However, the approaches named above tend to suffer from high computational complexity. In this work, we focus in particular on how neuro-symbolic AI can help to reduce the need of (labelled) data. To this end, we focus on parallel approaches which are more tailored to this problem as they enhance the neural model. However, both  T-S and DPL come with several limitations. 
Firstly, the adopted formulations do not support settings in which the inputs and the targets abide by more complex relations. In particular, they only support rules expressing constraints directly relating the input with the output data and do not support recursive formulas. However, in practical scenarios, the relationships between the inputs and the outputs may be modelled only via richer logical formulas that reference \textit{latent} information, i.e., information that is not available either in the input or output data. 
Additionally, with regards to DPL, the integration is based on the assumption of independence between the two components, which generally does not hold as they both depend upon the same input data.
Finally, DPL is bound to Markov Logic Networks (MLNs) \cite{MLN} limiting its applicability to classification only. 

\textbf{Contribution}. We present Concordia\footnote{Available on https://github.com/jonathanfeldstein/Concordia}, a parallel neurosymbolic framework that overcomes the above limitations.
Concordia adopts probabilistic logics due to their flexibility to reason in a formal fashion over uncertain data \cite{DeRaedt2015}.  
Concordia relies on the theoretically sound inference and training techniques of probabilistic logics to train the two components in a supervised or unsupervised fashion.
Its interface supports theories expressed as weighted formulas in first-order logic, including lifted graphical models like MLNs and Probabilistic Soft Logic (PSL) \cite{psl-long} allowing for easy integration of domain expertise.
Offering a plug-and-play interface is crucial as each language comes with its own 
semantics and inference/learning footprint.  
Furthermore, our framework employs a mixture of experts technique \cite{jacobs1991adaptive} to integrate the two components without relying on the independence assumption, 
and it supports using the deep network predictions as priors. Our empirical findings support the hypothesis that integrating symbolic models with neural ones does indeed reduce the need for (labelled) data.
Finally, as we empirically demonstrate,
learning the weights of the formulas at training time provides  
the means to learn the formulas of the theories themselves and not just their weights. In fact, we show that if we throw arbitrary formulas at training time, then the weights of the non-useful ones will drop to zero.
 
\textbf{Results} Concordia is the first framework of its kind to be applied beyond NLP and, in particular, on formulas that prior art in parallel approaches are incapable of supporting. 
Our evaluation shows that even simple commonsense formulas can have a significant impact on the accuracy of the end-task. In particular, 
Concordia leads to NLP models with up to 5.8\% higher accuracy than those of DPL
and to activity recognition models with up to 6.75\% and 3.87\% higher accuracy than the neurosymbolic techniques from \cite{collective-psl} and \cite{arxiv2020}, respectively. It also improves the root mean squared error on recommendation \cite{kouki2015hyper} by up to 2.10\%. These improvements increase even further with less data.
\section{Related work} \label{section:related}

\textbf{Stratified integration} 
As reported in \cite{tsamoura2020neuralsymbolic}, 
most existing stratified frameworks suffer from limited scalablity especially in the presence of recursion. Furthermore, the ones that employ heuristics to reduce the reasoning overhead, such as ABL \cite{abl}, suffer from limited accuracy. 
The above limitations restrict the applicability of frameworks such as \cite{deepproblog, abl, neurasp}
to real-world settings.
Concordia supports theories that relevant prior art fails to support, see Section~\ref{section:experiments} for further details.            

\textbf{Parallel integration} 
Above, we focused on several limitations of T-S and DPL, including inability to express complex relationships that govern the input and the output data. Hence, their applicability is restricted to simple tasks. 
With regards to T-S, there is an additional limitation: the optimization objective that is used to build the teacher model does not abide by the semantics of Probabilistic Soft Logic \cite{psl-long}. In particular, in the optimal solution found, when building a teacher, the slack variables (the $\xi$'s in \cite{hu-etal-2016-harnessing}) should approach 0 so that the expectation of the rule satisfaction becomes 1. However, in that case, the weight of each constraint is not taken into account.

\textbf{Knowledge distillation} Our work is relevant to the broader area of knowledge distillation \cite{distillation1, distillation2, distillation3}. There, 
the objective is to distill knowledge from a complex deep model into a simpler one. 
Our work substantially differs from the above line of research.
\textbf{(i)} Our teacher is a (probabilistic) logical theory and not a neural model. 
Therefore, Concordia allows integrating into a deep model 
prior knowledge in symbolic form, making introduction of domain expertise very simple.
\textbf{(ii)} At a higher level, 
prior art on purely neural teacher-student techniques  
aims to simplify a complex deep model. In contrast, we aim to improve the accuracy of a neural model using prior knowledge. 
\textbf{(iii)} In the above line of research, the accuracy of the student will not outperform that of the teacher. In contrast, our analysis shows our symbolic teacher can lead to 
neural models with higher accuracy than on its own. 

\textbf{Regularization}
LTNs \cite{LTN}, DL \cite{KR2020-92} and DASL \cite{DASL} introduce stratified techniques for training deep models using the semantics of fuzzy logic, however, without training the logic theory. 
pLogicNet \cite{pLogicNet} applies Markov Logic Networks to learn knowledge graph embeddings. Xie et al. \yrcite{DBLP:conf/nips/XieXMKS19} propose a technique to project fixed propositional (variable-free) formulas onto a manifold via graph
convolutional networks. The weights of the formulas are not jointly trained with the deep model. In \cite{dl2} and in \cite{li2019augmenting}, the authors propose frameworks where the neural models are regularised using symbolic constraints, by asserting whether the neural model satisfies \textit{one fixed} constraint. The above means that their loss is a SAT function and the constraints are unweighted. 

Beyond the difference about supporting theories rather than single constraints, Concordia distills the knowledge captured in the full probability distribution, rather than just the max of the logical component ($\logical$) into the neural component ($\neural$). Hence, we use more of the knowledge produced by $\logical$ -- our loss is a comparison of distributions. The experiments in Section~\ref{section:experiments} show that this additional knowledge is particularly helpful in settings with limited data.

\section{Preliminaries} \label{section:preliminaries}

\textbf{First-order logic}  
A \emph{term} is either a \textit{variable} or a \textit{constant}. An \textit{atom} $\alpha$ is an expression of the form $p(\vec{t})$, where $p$ is a \textit{predicate} and $\vec{t}$ is a vector of terms; $\alpha$ 
is \notion{ground}, if $\vec{t}$ includes only constants. 
We refer to expressions in first-order logic as \notion{formulas}. 
\notion{Rules} are universally quantified formulas of the form 
${\alpha_1 \wedge \dots \wedge \alpha_N \rightarrow \alpha}$, 
where $\alpha_i$'s and $\alpha$ are atoms and each term occurring in 
$\alpha$ also occurs in some $\alpha_i$. 
We refer to the left-hand and right-hand sides of the implication as the \textit{premise} and the \textit{conclusion} of the rule, respectively. 
A formula is \notion{instantiated} if each atom occurring in the formula is ground. A \notion{theory} is a collection of formulas.  
Assuming a set of constants, we refer to the set of all possible ground atoms computed by instantiating the formulas in the theory as its \textit{Herbrand base}. 
An \textit{interpretation} is a mapping of the ground atoms in the Herbrand base to a truth value. In the classical Boolean logic, each ground atom is mapped to true or false. Other logic formalisms map ground atoms to ${[0,1]}$. 

\begin{example}\label{ex:FOL}
{Consider the rule from \cite{kouki2015hyper}:
\begin{align}
\textsc{Similar}(I_1,I_2) \wedge \textsc{Rates}(U,I_1) \rightarrow \textsc{Rates}(U,I_2) \label{ex:FOL:rule}
\end{align}
According to the rule, if items $I_1$ and $I_2$ are similar, then user $U$ will rate item $I_1$. In this example, $I_1$, $I_2$, and $U$ are variables. 
Neither T-S nor DPL can support rule \eqref{ex:FOL:rule:grounding} due to their inability to handle transitive rules. 
An instantiation of rule \eqref{ex:FOL:rule:grounding}, where \texttt{T1}, \texttt{T2}, and \texttt{A} stand for \texttt{ToyStory1}, \texttt{ToyStory2}, and \texttt{Alice}, respectively, could be 
\begin{align}
\textsc{Similar}(\texttt{T1},\texttt{T2}) \wedge \textsc{Rates}(\texttt{A},\texttt{T1}) \rightarrow \textsc{Rates}(\texttt{A}, \texttt{T2}) \label{ex:FOL:rule:grounding}
\end{align}}
\end{example}

\textbf{Mappings} A \emph{substitution} $\sigma$ is a total mapping from variables to constants. 
For a vector $\vec{X}$, $\sigma(\vec{X})$ is obtained by replacing each occurrence of $X \in \vec{X}$ in the domain of $\sigma$ with $\sigma(X)$. 

\textbf{Markov Random Fields} A \emph{Markov Random Field} (MRF) is a model for the joint distribution of a set of {random} variables {(RVs)} ${\vec{X} = (X_1,\dots, X_N)}$. It is composed of an undirected graph $G$ and a set of \emph{potentials} ${\phi_1,\dots,\phi_M}$,
where each $\phi_i$ is usually represented by an indicator function $f_i$ weighted by $\lambda_i$. 
Graph $G$ has a node for each variable, while the MRF has a potential $\phi_i$ for each clique of variables 
$\vec{X}_i$ in $G$.
Potential $\phi_i$ maps each instantiation of $\vec{X}_i$ to a non-negative real value. The joint distribution represented by an MRF is given by:
\begin{align}\label{eq:log-linear}
    P({\vec{X} = \vec{x}}) = \frac{1}{Z} \exp \left( \sum_{i=1}^{M} \lambda_i f_i(\vec{X}_i = \vec{x}_i) \right),
\end{align}
where $Z$ is a normalization constant and $\vec{x}_i$ denotes an instantiation of  
$\vec{X}_i$ as per $\vec{x}$, i.e., assuming $\vec{x} = \sigma(\vec{X})$, for a substitution $\sigma$, then ${\vec{x}_i = \sigma(\vec{X}_i)}$. 
{We will see examples of $f_i$'s later in the section.} 

We denote the marginal distribution of the subset $\vec{Y}$ of $\vec{X}$ by $MARG(\vec{Y} = \vec{y})$ and compute 
conditional probabilities using the Bayes rule.
Let $\vec{X}^o$ denote a tuple of \textit{observed} RVs, i.e., RVs with known values, 
and $\vec{X}^u$ denoted a tuple of \textit{unobserved} RVs, i.e., RVs with unknown values.
The \notion{Most Probable Explanation} (MPE) or \notion{Maximum A Posteriori State} (MAP)
is the most likely assignment to the variables in $\vec{X}^u$ given $\vec{X}^o$: 
\begin{align}
    \begin{split}
        &MPE(\vec{X}^u = \vec{x}^u \mid \vec{X}^o = \vec{x}^o)\\ &= \argmax_{\vec{x}^u} P(\vec{X}^u = \vec{x}^u \mid \vec{X}^o = \vec{x}^o)
    \end{split}
\end{align} 

\textbf{LGMs} To establish the connection between a theory in first order logic $\logical$ and probability theory, \emph{lifted graphical models} (LGMs) treat each ground atom in the Herbrand base of $\logical$ as a random variable (RV) with the same domain as the one of the atom. 
Each formula $r_i$ in $\logical$ serves as a template for defining potentials $\phi_{i,1}, \dots, \phi_{i,M_i}$. Those potentials share the same weights $\lambda_i$ (as we shall later see, this is going to be the weight of $r_i$) and map each tuple of constants $\vec{x}$ to the same value. To summarize, 
an LGM is a set of weighted formulas ${\lambda_i::r_i}$, for ${1 \leq i \leq M}$, where each $r_i$ defines a set of potentials $\phi_{i,1},\dots,\phi_{i,M_i}$ in a MRF with $\phi_{i,j} = \phi_{i,k}$, for each $j,k \in \{1, \dots, M_i\}$. The log-linear representation of LGMs defines the joint distribution
\begin{align*}
    P({\vec{X} = \vec{x}}) = \frac{1}{Z} \exp \left( \sum_{i=1}^{M} \sum_{j=1}^{M_{i}} \lambda_i f_{i,j}(\vec{X}_{i,j} = \vec{x}_{i,j}) \right),
\end{align*}
where $Z$ is the normalization constant and $\vec{x}_{i,j}$ is defined analogously to \eqref{eq:log-linear}. 

\textit{Probabilistic Soft Logic} (PSL) \cite{psl-long} is an example of an LGM. PSL adopts the semantics of fuzzy logic to interpret the formulas and the Lukasiewicz t-(co)norms to compute the truth values of the instantiated formulas. 
Due to the adoption of fuzzy logic, interpretations map atoms to soft truth values in ${[0,1]}$.
Returning back to Example~\ref{ex:FOL}, the truth value of atom 
\textsc{Similar}(\texttt{ToyStory1},\texttt{ToyStory2}) is in $[0,1]$ in PSL. That allows us to incorporate uncertainty to the level of similarity between elements, something that is not possible with Boolean first-order logic.
The $f_{i,j}$ functions are given by 
\begin{align} \label{eq:pslfactor}
    f_{i,j}(\vec{X}_{i,j} = \vec{x}_{i,j}) = (1 -r_i(\vec{X}_{i,j} =\vec{x}_{i,j}))^p
\end{align}
where ${r_i(\vec{x}_{i,j})}$ denotes the truth value of 
formula $r_i$ when instantiated using the ground atoms $\vec{x}_{i,j}$ and 
${p \in \{1,2\}}$ provides a choice of penalty.
If rule $r_i$ cannot be instantiated using $\vec{x}_{i,j}$, then 
${r_i(\vec{x}_{i,j}) = 0}$.

\section{Combining logic with neural networks}\label{section:core}
% 
%%%%%Concordia FIGURE
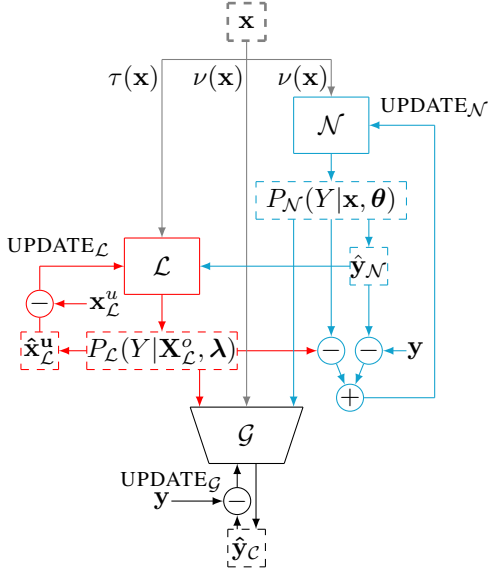
\begin{figure}[!htb]
    \centering
    \begin{tikzpicture}
	\begin{pgfonlayer}{nodelayer}
		\node [style=none] (0) at (0, 4.375) {$\mathbf{x}$};
		\node [style=none] (1) at (-0.25, 4.625) {};
		\node [style=none] (2) at (0.25, 4.625) {};
		\node [style=none] (3) at (0.25, 4.125) {};
		\node [style=none] (4) at (-0.25, 4.125) {};
		\node [style=none] (5) at (0, 4.125) {};
		\node [style=none] (7) at (1.125, 3.875) {};
		\node [style=none] (8) at (-1.125, 3.875) {};
		\node [style=none] (9) at (1.125, 3.375) {};
		\node [style=none] (10) at (1.625, 3.375) {};
		\node [style=none] (11) at (0.625, 3.375) {};
		\node [style=none] (12) at (1.625, 2.625) {};
		\node [style=none] (13) at (0.625, 2.625) {};
		\node [style=none] (14) at (1.125, 3) {$\neural$};
		\node [style=none] (15) at (1.625, 3) {};
		\node [style=none] (16) at (1.125, 2.25) {};
		\node [style=none] (17) at (1.125, 2.625) {};
		\node [style=none] (18) at (0.125, 2.25) {};
		\node [style=none] (19) at (0.125, 1.75) {};
		\node [style=none] (20) at (2.125, 1.75) {};
		\node [style=none] (21) at (2.125, 2.25) {};
		\node [style=none] (22) at (1.125, 2) {$\probneural(Y | \mathbf{x}, \nparams)$};
		\node [style=none] (23) at (1.125, 1.75) {};
		\node [style=small pink] (27) at (1.125, 0) {$-$};
		\node [style=none] (28) at (1.5, 1.125) {};
		\node [style=none] (29) at (1.375, 0.875) {};
		\node [style=none] (30) at (1.375, 1.375) {};
		\node [style=none] (31) at (1.875, 1.375) {};
		\node [style=none] (32) at (1.875, 0.875) {};
		\node [style=none] (33) at (1.625, 1.125) {$\hat{\vec{y}}_{\neural}$};
		\node [style=small pink] (34) at (1.625, 0) {$-$};
		\node [style=none] (35) at (1.625, 0.875) {};
		\node [style=none] (36) at (2.125, 0) {};
		\node [style=small pink] (37) at (1.375, -0.625) {$+$};
		\node [style=none] (38) at (2.5, -0.625) {};
		\node [style=none] (39) at (2.5, 3) {};
		\node [style=none] (40) at (-1.125, 1.125) {$\logical$};
		\node [style=none] (41) at (-1.625, 1.5) {};
		\node [style=none] (42) at (-1.625, 0.75) {};
		\node [style=none] (43) at (-0.625, 1.5) {};
		\node [style=none] (44) at (-0.625, 0.75) {};
		\node [style=none] (46) at (-2.125, 0.25) {};
		\node [style=none] (47) at (-2.125, -0.25) {};
		\node [style=none] (48) at (-0.125, -0.25) {};
		\node [style=none] (49) at (-0.125, 0.25) {};
		\node [style=none] (50) at (-1.125, 0) {${\problogical(Y | \lveco, \lparams)}$};
		\node [style=none] (51) at (-1.125, 0.75) {};
		\node [style=none] (52) at (-1.125, 0.25) {};
		\node [style=none] (53) at (1.375, 1.125) {};
		\node [style=none] (54) at (-0.625, 1.125) {};
		\node [style=none] (57) at (-1.125, 1.5) {};
		\node [style=none] (58) at (-0.125, 0) {};
		\node [style=none] (60) at (-2.875, -0.25) {};
		\node [style=none] (61) at (-3, 0.25) {};
		\node [style=none] (62) at (-2.5, 0.25) {};
		\node [style=none] (63) at (-2.5, -0.25) {};
		\node [style=none] (64) at (-3, -0.25) {};
		\node [style=none] (65) at (-2.75, 1.125) {};
		\node [style=smallred] (66) at (-2.75, 0.625) {$-$};
		\node [style=none] (67) at (-2.75, 0.25) {};
		\node [style=none] (68) at (-2.75, 0) {$\mathbf{\hat{x}^u_{\logical}}$};
		\node [style=none] (69) at (-1.625, 1.125) {};
		\node [style=none] (70) at (-2.125, 0.625) {};
		\node [style=none] (71) at (-1.875, 0.625) {$\mathbf{x}^u_{\logical}$};
		\node [style=none] (72) at (0, -0.75) {};
		\node [style=none] (73) at (0.625, -0.75) {};
		\node [style=none] (74) at (-0.625, -0.75) {};
		\node [style=none] (76) at (-0.625, -0.25) {};
		\node [style=none] (77) at (0.75, -0.75) {};
		\node [style=none] (78) at (-0.75, -0.75) {};
		\node [style=none] (79) at (0.5, -1.5) {};
		\node [style=none] (80) at (-0.5, -1.5) {};
		\node [style=none] (81) at (0, -1.125) {$\gating$};
		\node [style=none] (82) at (0.125, -1.5) {};
		\node [style=none] (83) at (-0.125, -1.5) {};
		\node [style=none] (84) at (0.125, -2.375) {};
		\node [style=none] (85) at (-0.125, -2.375) {};
		\node [style=smallCircle] (86) at (-0.125, -2) {$-$};
		\node [style=none] (87) at (-1, -2) {};
		\node [style=none] (88) at (-1.125, -2) {$\vec{y}$};
		\node [style=none] (89) at (-0.25, -2.375) {};
		\node [style=none] (90) at (-0.25, -2.875) {};
		\node [style=none] (91) at (0.25, -2.875) {};
		\node [style=none] (92) at (0.25, -2.375) {};
		\node [style=none] (93) at (0, -2.625) {$\vec{\hat{y}}_{\concordia}$};
		\node [style=none] (94) at (0.625, 1.75) {};
		\node [style=none] (95) at (1.625, 1.75) {};
		\node [style=none] (96) at (1.625, 1.375) {};
		\node [style=none] (97) at (2.25, 0) {$\vec{y}$};
		\node [style=none] (98) at (-2.125, 0) {};
		\node [style=none] (99) at (-2.5, 0) {};
		\node [style=none] (100) at (-2.5, 1.375) {$\update_{\logical}$};
		\node [style=none] (101) at (2.5, 3.25) {$\update_{\neural}$};
		\node [style=none] (102) at (-1, -1.75) {$\update_{\gating}$};
		\node [style=none] (103) at (-1.5, 3.625) {$\tau(\mathbf{x})$};
		\node [style=none] (104) at (0.75, 3.625) {$\nu(\mathbf{x})$};
		\node [style=none] (105) at (-0.375, 3.625) {$\nu(\mathbf{x})$};
	\end{pgfonlayer}
	\begin{pgfonlayer}{edgelayer}
		\draw [style=undirected dashed] (1.center) to (2.center);
		\draw [style=undirected dashed] (3.center) to (4.center);
		\draw [style=undirected dashed] (4.center) to (1.center);
		\draw [style=undirected] (7.center) to (8.center);
		\draw [style=pinksolid] (10.center) to (11.center);
		\draw [style=pinksolid] (12.center) to (13.center);
		\draw [style=pinksolid] (10.center) to (12.center);
		\draw [style=pinksolid] (11.center) to (13.center);
		\draw [style=pinkdirected] (17.center) to (16.center);
		\draw [style=pink dashed] (19.center) to (20.center);
		\draw [style=pink dashed] (19.center) to (18.center);
		\draw [style=pink dashed] (18.center) to (21.center);
		\draw [style=pink dashed] (21.center) to (20.center);
		\draw [style=pinkdirected] (23.center) to (27);
		\draw [style=directed] (7.center) to (9.center);
		\draw [style=pink dashed] (30.center) to (31.center);
		\draw [style=pink dashed] (31.center) to (32.center);
		\draw [style=pink dashed] (32.center) to (29.center);
		\draw [style=pink dashed] (29.center) to (30.center);
		\draw [style=pinkdirected] (35.center) to (34);
		\draw [style=pinkdirected] (36.center) to (34);
		\draw [style=pinkdirected] (27) to (37);
		\draw [style=pinkdirected] (34) to (37);
		\draw [style=pinksolid] (37) to (38.center);
		\draw [style=pinksolid] (38.center) to (39.center);
		\draw [style=pinkdirected] (39.center) to (15.center);
		\draw [style=undirected red] (41.center) to (43.center);
		\draw [style=undirected red] (43.center) to (44.center);
		\draw [style=undirected red] (44.center) to (42.center);
		\draw [style=undirected red] (42.center) to (41.center);
		\draw [style=reddashed] (47.center) to (48.center);
		\draw [style=reddashed] (47.center) to (46.center);
		\draw [style=reddashed] (46.center) to (49.center);
		\draw [style=reddashed] (49.center) to (48.center);
		\draw [style=directed red] (51.center) to (52.center);
		\draw [style=directed] (8.center) to (57.center);
		\draw [style=directed red] (58.center) to (27);
		\draw [style=reddashed] (61.center) to (64.center);
		\draw [style=reddashed] (64.center) to (63.center);
		\draw [style=reddashed] (63.center) to (62.center);
		\draw [style=reddashed] (62.center) to (61.center);
		\draw [style=directed red] (65.center) to (69.center);
		\draw [style=undirected red] (65.center) to (66);
		\draw [style=undirected red] (66) to (67.center);
		\draw [style=directed red] (70.center) to (66);
		\draw [style=directed red] (76.center) to (74.center);
		\draw (77.center) to (78.center);
		\draw (77.center) to (79.center);
		\draw (79.center) to (80.center);
		\draw (80.center) to (78.center);
		\draw [style=directed] (5.center) to (72.center);
		\draw [style=black] (86) to (83.center);
		\draw [style=black] (85.center) to (86);
		\draw [style=black] (82.center) to (84.center);
		\draw [style=black] (87.center) to (86);
		\draw [style=blackdashed] (90.center) to (91.center);
		\draw [style=blackdashed] (91.center) to (92.center);
		\draw [style=blackdashed] (92.center) to (89.center);
		\draw [style=blackdashed] (89.center) to (90.center);
		\draw [style=undirected dashed] (2.center) to (3.center);
		\draw [style=pinkdirected] (53.center) to (54.center);
		\draw [style=pinkdirected] (94.center) to (73.center);
		\draw [style=pinkdirected] (95.center) to (96.center);
		\draw [style=directed red] (98.center) to (99.center);
	\end{pgfonlayer}
\end{tikzpicture}
    \caption{Overview of the Concordia architecture. Grey components are the input, blue components are the neural component, red components are the logic component, and the components parts are the gating network.}
    \label{fig:concordia}
\end{figure}
This section presents the main contribution of this work, the Concordia framework. An overview of the framework is presented in Figure \ref{fig:concordia}.

Let $\trainingdata$ denote the set of training data.
Each training datum is denoted by 
${(\vec{x},\vec{y})}$, where $\vec{x} \in \mathcal{X}$ and $\vec{y} \in \mathcal{Y}$.
Concordia uses each ${(\vec{x},\vec{y})}$ to train the neural component, the logical theory, and a gating network, denoted by $\neural$, $\logical$ and $\gating$, respectively.
We first present the $\neural$ and $\logical$ components and then proceed with the description of the overall architecture. 

\subsection{The logical component}\label{section:core:logical}
The first building block of our architecture is an LGM $\logical$.
We denote the vector of all formulas' weights as $\lparams$. 
Treating each ground atom 
in the Herbrand base of $\logical$ as a RV, 
module $\logical$ defines a probability distribution 
$\problogical(\lvec | \lparams)$, see Section~\ref{section:preliminaries}.

Component $\logical$ should infer information for each possible target in $\mathcal{Y}$. This implies that the Herbrand base of $\logical$ needs to include a ground atom $\alpha_j$ denoting the truth of each target class $c_j \in \mathcal{Y}$. 
We refer to atoms $\alpha_j$ as \textit{target atoms}. 
If $\mathcal{Y}$ is a continuous space, as in regression tasks, the adoption of PSL
allows us to use rules so that the Herbrand base of $\logical$ includes a single ground atom per target.

The training data defines instantiations of the RVs in $\lvec$. 
In particular, ${\tau(\vec{x})}$ and ${\tau(\vec{x}\vec{y})}$ are vectors of 
(partial) truth assignments to the elements in $\lvec$.
Logic allows us to train under \textit{uncertain data}
by assigning values to the atoms in the range [0,1]. 
By abusing the notation, we can also treat the $\tau$-vectors as mappings of atoms to their truth values. 
We use $\lveco$ to denote the vectors of RVs that are instantiated (observed) given the vectors in 
${\tau(\vec{x})}$ or ${\tau(\vec{x}\vec{y})}$; we use 
$\lvecu$ to denote the RVs that are left non-instantiated.

\begin{example} \label{ex:logicComponent}
    Reconsider rule~\eqref{ex:FOL:rule}, where we now use $\lambda$ to denote its weight.
    Each datum $(\vec{x}, \vec{y})$ is a user-item pair.
    In particular, vector ${\tau(\vec{x}\vec{y})}$ includes the relational representation of users and items, e.g.,  
    if $(\vec{x}, \vec{y})$ is the rating, say 3, of user \texttt{Alice} on the movie \texttt{Toy Story}, then the vector ${\tau(\vec{x}\vec{y})}$ assigns 
    to atom ${\textsc{Rates}(\texttt{Alice}, \texttt{Toy Story})}$ the rating 3 normalized into ${[0,1]}$ (that is 0.5).
\end{example}

\textbf{Inference}
Inference proceeds by means of a conditional distribution ${\problogical(Y  | \lveco, \lparams)}$, where $Y$ is a RV with domain $\mathcal{Y}$,
denoting the likelihood of a target for given observations (instantiations to the observed variables $\lveco$) and formula weights $\lparams$. Continuing with Example \ref{ex:logicComponent}, assuming our goal is to predict the rating of 
    $\texttt{Alice}$ on $\texttt{Toy Story}$, then ${\textsc{Rates}(
    \texttt{Alice}, \texttt{Toy Story})}$ is the single target atom and ${\problogical(Y | \lveco, \lparams)}$ denotes the conditional distribution over the values of the target.

To define ${\problogical(Y | \lveco, \lparams)}$ in classification tasks, we have two options depending on the logic semantics.
When the logic admits Boolean interpretations, i.e., an atom is either true or false with some probability as in Markov Logic Networks \cite{MLN},
then we can compute the likelihood of each 
$\alpha_{j}$ being true using marginals and arrange the computed likelihoods into distributions over the classes.

In contrast, when the logic interpretations map each ground atom to ${[0,1]}$,
${\problogical(Y = c_{j} | \lveco = \tau(\vec{x}), \lparams)}$ is the soft truth value of the atom $\alpha_{j}$ in the interpretation
${\tau(\vec{x}) \lvecuasspred}$, where 
\begin{align}
    \lvecuasspred = \argmax_{\lvecuass} \problogical(\lvecu = \lvecuass | \lveco = \tau(\vec{x}), \lparams) 
\end{align}
Above, $\lvecuasspred$ denotes the most likely assignment of the unobserved variables given the input data and the current weights and ${\tau(\vec{x}) \lvecuasspred}$ denotes the assignment of the atoms in the Herbrand base of $\logical$ by taking the union of the assignments in $\tau(\vec{x})$ and $\lvecuasspred$. The discussion on inference for regression tasks is deferred to Appendix \ref{app:logical}.

\textbf{Constraints} When $\logical$ admits Boolean interpretations,
we require it to impose mutual exclusiveness constraints in the elements in $\mathcal{Y}$, i.e., that the elements in $\mathcal{Y}$ cannot be simultaneously true. 
When $\logical$ admits interpretations in ${[0,1]}$, then we require $\logical$ to impose the constraint that the sum of the soft truth values of the $\atom_{j}$'s in each interpretation equals to 1. 
Both constraints ensure that the distribution over classes is a valid one, i.e., the sum over all possible outcomes is 1.

\textbf{Training} 
Training aims to learn the weights $\lparams$.
The task is formalized as finding the $\lparams$ maximizing the log likelihood of the assignments $\tau(\vec{x}\vec{y})$:
\begin{align}
    \hat{\lparams} = \argmax_{\lparams} \prod \limits_{(\vec{x}, \vec{y}) \in \trainingdata} \problogical(\lvec = \tau(\vec{x} \vec{y}), \lparams) \label{eq:ltraining}
\end{align}
Equation \eqref{eq:ltraining} works when $\tau(\vec{x} \vec{y})$ provides truth assignments to all variables in 
$\lvec$. If that assumption is violated, training resorts to an expectation-maximization problem. 
Parameter learning in lifted graphical models
as well as probabilistic logic programs works via gradient ascent \cite{MLN, psl-long}.
We use 
${\update_\logical(\vec{x}, \vec{y}, \tau, \lparams_t) \rightarrow \lparams_{t+1}}$ to denote  
updating of the parameters in $\logical$ given ${(\vec{x}, \vec{y})}$.

\subsection{The neural component}\label{section:core:neural}
Given a neural model $\neural$, then, for a fixed assignment to the neural weights $\nparams$, 
$\neural$ defines a conditional distribution 
$\probneural(Y| \nvec, \nparams)$ by using a softmax output layer producing a 
${|\mathcal{Y}|}$-dimensional prediction vector. 
Then, $\probneural(Y = c_{j} | \nvec, \nparams)$
defines the likelihood the label is class $c_{j} \in \mathcal{Y}$.

Similarly to the previous section, we have to specify the semantics of the training data. 
We denote by $\nu(\vec{x})$ the vector representation of $\vec{x}$ in the input format of $\neural$. 

\textbf{Inference} For fixed $\nparams$ and $\nu(\vec{x})$, 
$\neural$ returns prediction ${\hat{\vec{y}}}$:
\begin{align}
    \hat{\vec{y}} = \argmax_{\vec{y}} \probneural(Y=\vec{y} | \nvec = \nu(\vec{x}), \nparams) \label{eq:neural:inference}
\end{align}
We use  
${\infer_\neural(\vec{x}, \nu, \nparams) \rightarrow \hat{\vec{y}}}$ to denote inference in $\neural$.

\textbf{Training} 
Training is achieved by backpropagation, as standard in deep networks,
via a loss function measuring the discrepancy between the neural predictions and the true label.

\subsection{Integration of components}\label{section:core:concordia}
The integration of the two components in Concordia is based on a gating network $\gating$
with learnable parameters $\gparams$. In our implementation, the input domain of $\gating$ is the same as of $\neural$. The output domain is ${[0,1]}$. The idea is that depending on the data the benefit provided by either model can vary. Prior art discards the logical model entirely at the end of the training, or simply multiplies the two distributions under the assumption that the two distributions are independent. Our approach is more flexible and takes advantage of the logical model during testing.

\paragraph{Inference}  
Inference is described in Algorithm~\ref{algorithm:inference}.
For each input datatum 
$\vec{x}$, we first provide $\nu(\vec{x})$ and $\tau(\vec{x})$ to $\neural$
and $\logical$, to get the 
distributions $\probneural(Y| \nvec = \nu(\vec{x}), \nparams)$
and ${\problogical(Y | \lveco = \tau(\vec{x}), \lparams)}$. Then, we combine those distributions using 
$\gating$. 
In particular, Concordia defines the following conditional probability distribution:
\begin{align}
\begin{split}
    &\probconcordia(Y | \nvec, \lveco, \nparams, \lparams, \gparams)\\ &= \gating(\nvec, \gparams) \probneural(Y| \nvec, \nparams)\\  &+ 
    (1-\gating(\nvec, \gparams)) \problogical(Y| \lveco, \lparams) \label{eq:mixture}
\end{split}
\end{align}
Prediction $\hat{\vec{y}}$ given input $\vec{x}$ and $\nparams$, $\lparams$, $\gparams$ is then computed as
\begin{align}
    \hat{\vec{y}} = \argmax_{\vec{y}} \probconcordia(Y = \vec{y} |\nvec = \nu(\vec{x}), \lveco = \tau(\vec{x}), \nparams, \lparams, \gparams) \label{eq:concordia:prediction}
\end{align}    
Notice that above formulation is substantially different from the ones proposed in T-S and DPL: T-S uses posterior regularization; DPL is based on multiplying the predictions. 

Forward inference in Concordia is denoted via
${\infer_\concordia(\vec{x}, \nu, \tau, \nparams, \lparams, \gparams) \rightarrow \hat{\vec{y}}}$.

\begin{minipage}[tb]{\columnwidth}
\begin{algorithm}[H]
   \caption{$\infer_\concordia$($\vec{x}, \nu, \tau, \nparams, \lparams, \gparams$)} \label{algorithm:inference}
    \begin{algorithmic}
        \STATE ${\Delta_{\neural}(Y) \leftarrow \probneural(Y| \nvec = \nu(\vec{x}), \nparams)}$
        \STATE ${\Delta_{\logical}(Y) \leftarrow \problogical(Y| \lveco = \tau(\vec{x}), \lparams)}$
        \STATE $\kappa \leftarrow \gating(\nvec, \gparams)$
        \STATE ${\Delta_{\concordia}(Y) \leftarrow \kappa \Delta_{\neural}(Y) + (1-\kappa) \Delta_{\logical}(Y)}$
        \STATE \textbf{return} ${\argmax_{\vec{y}} \Delta_{\concordia}(Y = \vec{y})}$
    \end{algorithmic}
\end{algorithm}
\end{minipage} 
\paragraph{Training} 
Parameter update in Concordia is summarized in Algorithm~\ref{algorithm:training}.
For each ${(\vec{x}, \vec{y})}$ training proceeds as follows.
Firstly, $\logical$ is trained based on $(\vec{x}, \vec{y})$ and the interface 
$\update_\logical$ as described in Section~\ref{section:core:logical}.  
Secondly, $\neural$ is trained to minimize the difference between its predictions and the true labels $\vec{y}$, as well as to minimize the difference between distributions $\probneural(Y| \nvec = \nu(\vec{x}), \nparams)$
and $\problogical(Y| \lveco = \tau(\vec{x}), \lparams)$. 
The first term aims to keep $\neural$'s predictions close to the ground truth. 
The second term aims to keep $\neural$'s predictions close to the ones of $\logical$. Hence $\logical$ supervises $\neural$ in a weak fashion during the training process. The neural weights get updated at each step $t$ as follows: 
\begin{align}
\begin{split}
    &\nparams_{t+1} = \argmin_{\nparams} \ell(\hat{\vec{y}}, \vec{y}) +\\ &
    KL(\probneural(Y| \nvec = \nu(\vec{x}), \nparams_{t})||\problogical(Y| \lveco = \tau(\vec{x}), \lparams_t)) \label{eq:training:neural}
\end{split}
\end{align}
Above, ${\hat{\vec{y}} = \infer_\neural(\vec{x}, \nu, \nparams)}$
and 
$KL$ denotes the Kullback–Leibler divergence between distributions.
The $\neural$'s parameter update process we described above is denoted via
${\update_\neural(\vec{x}, \vec{y}, \nu, \nparams_t) \rightarrow \nparams_{t+1}}$.

Lastly, parameters $\gparams$ of $\gating$ are amended so that the labels predicted by the whole framework 
${\hat{\vec{y}}_{\concordia}}$
fit the ground truth: 
\begin{align}
    \gparams_{t+1} &= \argmin_{\gparams} \ell(\vec{y}, \hat{\vec{y}}_{\concordia}) \label{eq:training:gating}
\end{align}
Above, ${\hat{\vec{y}}_{\concordia} = \infer_\concordia(\vec{x}, \nu, \tau, \nparams, \lparams, \gparams)}$.

\begin{minipage}[tb]{\columnwidth}
\begin{algorithm}[H]
   \caption{$\update_\concordia$($\vec{x}, \vec{y}, \nu, \tau, \nparams_t, \lparams_t, \gparams_t$) } \label{algorithm:training}
    \begin{algorithmic}
    \STATE $\nparams_{t+1} \leftarrow \update_{\neural}(\vec{x}, \vec{y}, \nu, \nparams_t)$
    \STATE $\lparams_{t+1} \leftarrow \update_{\logical}(\vec{x}, \vec{y}, \tau, \lparams_t)$
    \STATE ${\hat{\vec{y}}_{\concordia} \leftarrow \infer_{\concordia} (\vec{x}, \nparams_t, \lparams_t, \gparams_t)}$
    \STATE ${\gparams_{t+1} = \argmin_{\gparams} \ell(\vec{y}, \hat{\vec{y}}_{\concordia})}$
    \STATE \textbf{return} $(\nparams_{t+1}, \lparams_{t+1}, \gparams_{t+1})$
    \end{algorithmic}
\end{algorithm}
\end{minipage}

\textbf{Unsupervised learning} So far, we discussed supervised learning. 
Unsupervised learning is supported by discarding the first term in \eqref{eq:training:neural} and the gating network. Notice that in the unsupervised setting, we do not train $\logical$. Training of $\logical$ in an unsupervised setting is left for future work.  

\textbf{Multitasking} Above, the explanations focused on a single target atom. However, the architecture is easily extendable to multitasking. To this end, we consider instead of a single RV $Y$ a vector of RVs $\vec{Y}$ and the probability of $\neural$ becomes $\probneural(\vec{Y}=\vec{y} | \nvec = \nu(\vec{x}), \nparams)$. This allows on one hand to achieve different tasks in parallel and on the other to optimize predictions concurrently. For example, in the case of recommendation above, users and items are dependent on each other and need to be optimized concurrently. Multitasking is not supported by prior art. 

\subsection{Neural predictions as priors}
\label{section:core:priors}

We can use the neural predictions as weak supervision signal to train the logical component. 
Adding neural priors can be achieved by considering the neural predictions as additional observed RVs in $\logical$. 
The above is always possible via two steps: 
introducing additional rules propagating information from $\neural$ to $\logical$, and assigning to the atoms denoting the neural predictions the confidences predicted by $\neural$.

\begin{example}
\label{example:priors}
We demonstrate propagating information from 
$\neural$ to $\logical$ we extend Example \ref{ex:logicComponent}: 
{To introduce the rating predictions of $\neural$ as priors to $\logical$, we firstly add 
\begin{align}
   \lambda: \textsc{dnn}(U,I) \rightarrow \textsc{Rates}(U,I) \nonumber
\end{align} 
Atom $\textsc{dnn}(U,I)$ represents ratings of user $U$ on item $I$ as predicted by the deep network.} 
\end{example}

Algorithm~\ref{algorithm:translate} formalizes the creation of variable instantiations based on the neural predictions. 
After adding additional rules to $\logical$ as in Example~\ref{example:priors}, 
its Herbrand base includes a ground atom $\textsc{DNN}_{j}$ associated with the 
likelihood to which $\neural$ predicts the class $c_{j}$.  
We refer to $\textsc{DNN}{j}$ as the \textit{neural atom} for $c_{j}$ in the Herbrand base of $\mathcal{L}$. 
Algorithm~\ref{algorithm:translate} simply assigns to $\textsc{DNN}{j}$  
the likelihood $\probneural(Y = c_{j}| \nvec = \nu(\vec{x}), \nparams)$. 

To incorporate neural priors in Concordia's inference or learning process, we simply need to update the instantiations of $\lveco$ from 
$\tau(\vec{x})$ or $\tau(\vec{x} \vec{y})$
to $\tau(\vec{x}) \vec{z}$ and $\tau(\vec{x} \vec{y}) \vec{z}$,
where $\vec{z}$ denotes the computed instantiations (Alg. ~\ref{algorithm:translate}). 

\begin{algorithm}[tb]
   \caption{\translate($\vec{x}, \nu, \nparams, \logical$) $\rightarrow$ $\vec{z}$} \label{algorithm:translate}
    \begin{algorithmic}
    \STATE $\vec{z} \leftarrow \emptyset$ \COMMENT{$\vec{z}$ maps atoms to their truth values.}
        
        \FOR{\textbf{each} $c_{j} \in \mathcal{Y}$}
            \STATE $\textsc{DNN}_{j}$ denotes the neural atom for $c_{j}$ in the Herbrand base of 
            $\logical$ 
            \STATE \textbf{set} the value of $\textsc{DNN}_{j}$ in $\vec{z}$ to $\probneural(Y = c_{j}| \nvec = \nu(\vec{x}), \nparams)$ 
    
    \ENDFOR
    \STATE \textbf{return} $\vec{z}$
    \end{algorithmic}
\end{algorithm}

\section{Experiments}\label{section:experiments}

\textbf{Scenarios} We consider scenarios that have been adopted by prior symbolic and neurosymbolic techniques: {collective activity detection} (CAD) \cite{CVPR19, arxiv2020, collective-psl}, recommendation \cite{kouki2015hyper} and entity linking \cite{DBLP:conf/emnlp/WangP18}. 
To ensure a fair comparison, we adopt the same datasets and baselines with 
those techniques for each case.
Each scenario includes a neural component $\neural$ and a logical component $\logical$. $\logical$ is implemented in PSL for each scenario.
We denote by $\concordia(\neural, \logical)$ the Concordia instantiations. Further details on the experimental setup and training parameters can be found in Appendix \ref{app:experiments}.

\textbf{Baselines} We consider the following baselines:
\begin{itemize}[leftmargin=*]    
    \item \textbf{Recommendation.} We compare against (i) the deep models  
    \textbf{\nnmf}~ \cite{dziugaite2015neural}, \textbf{\neumf}~ \cite{he2017neural} and \textbf{\texttt{GraphRec}} \cite{graphrec}; and (ii) the purely symbolic technique from \cite{kouki2015hyper}, which performs the task using PSL. This dataset is used to test Concordia on regression tasks.
    \item \textbf{CAD.} We compare against 
    (i) the state-of-the-art models from \cite{CVPR19}
    that use \textbf{\mobilenet~}and \textbf{\inception~} as backbone networks;
    (ii) the state-of-the-art neurosymbolic technique {\textbf{\IARG}}\footnote{The authors adopt a Graph Convolutional Network (GCN) and extend on \cite{CVPR19} with a more sophisticated message passing algorithm.} \cite{arxiv2020} that is based on the above deep architectures, denoted as \IARG(\mobilenet) and \IARG(\inception), respectively; and
    (iii) the stratified neurosymbolic technique from \cite{collective-psl}. This dataset is used to test Concordia on complex classification.
    
    \item \textbf{Entity linking.} We adopt the same experimental setting with \cite{DBLP:conf/emnlp/WangP18} 
    and compare against (i) \textbf{\bilstm}, a Bi-LSTM recurrent neural network proposed in 
    \cite{peng2017cross}; (ii) \textbf{\dpl}; (iii) the symbolic theory adopted by the $\dpl$ authors alone; and (iv) \textbf{\distilbert} \cite{distilbert} a distilled version of BERT, a pre-trained transformer model originally proposed in \cite{bert}. Below, we denote \distilbert~ by \bert~ for readability purposes. This dataset is used to have a direct comparison with the closest competitor in the literature, as well as testing Concordia's performance in unsupervised and semi-supervised settings. 
\end{itemize}
\textbf{Other neuro-symbolic comparisons} Neither T-S nor DPL supports the theories adopted in CAD, due to latent variables, and recommendations, due to recursiveness, and, hence, they are not considered in those scenarios. 
In the appendix, we provide further details on the rules that are not supported by T-S and DPL. 

Furthermore, \textbf{TensorLog} \cite{DBLP:journals/jair/CohenYM20} only supports rules of a specific transitive form and therefore cannot be used in our experiments, while techniques such as Neural Theorem Proving \cite{DBLP:journals/corr/abs-1807-08204} are not relevant as they do not concern knowledge distillation, but question answering in Prolog.
Based on discussions with the authors of \textbf{DeepProbLog} \cite{deepproblog}, a comparison cannot be performed either: firstly, facts cannot be updated in ProbLog, secondly, it does not support learning across time sequences. Both limitations make DeepProbLog inapplicable to our scenarios. 
Additionally, the recommendation task has a very large number of entities in the knowledge base that can increase the training overhead of DeepProbLog to the extend that training is prohibitively slow (DeepProbLog faces significant scalability restrictions as observed in \cite{tsamoura2020neuralsymbolic}).
Neurosymbolic frameworks like \textbf{NeuroLog} \cite{tsamoura2020neuralsymbolic} and \textbf{ABL} \cite{abl} that rely on the same principles with DeepProbLog cannot be applied for the same reasons. 
Finally, regarding \textbf{DL2} \cite{dl2} and \cite{li2019augmenting}, beyond the limitations mentioned in Section~\ref{section:related} including the fact that these frameworks only support a single fixed constraint, they do not support passing latent variables. Furthermore, \cite{li2019augmenting} expects the rules to be of a specific acyclic form. For the above reasons, these two frameworks  do not support the experiments in this section. 

\textbf{Learning paradigms.}
Learning proceeds in a supervised fashion in the first two scenarios and in an unsupervised and semi-supervised fashion in the last one.
To assess the robustness of Concordia to the input theory, we consider a large set of noisy rules (i.e., rules that do not contribute to the task) with unknown weights and learn the weights of the rules at training time, in the last scenario.

\subsection{Item recommendation}

Given a user and his ratings, we aim to 
determine the user's prospective rating on an item.
We used the 2020 Yelp and MovieLens-100k datasets \cite{10.1145/2827872}
and ran the setups $\concordia(\nnmf, \theoryrec)$, $\concordia(\neumf, \theoryrec)$ as well as $\concordia(\texttt{GraphRec}, \theoryrec)$ on MovieLens-100k, where 
$\theoryrec$ is the PSL theory from \cite{kouki2015hyper}.

\begin{figure}[ht]
    \centering
    \includegraphics[width=0.95\columnwidth]{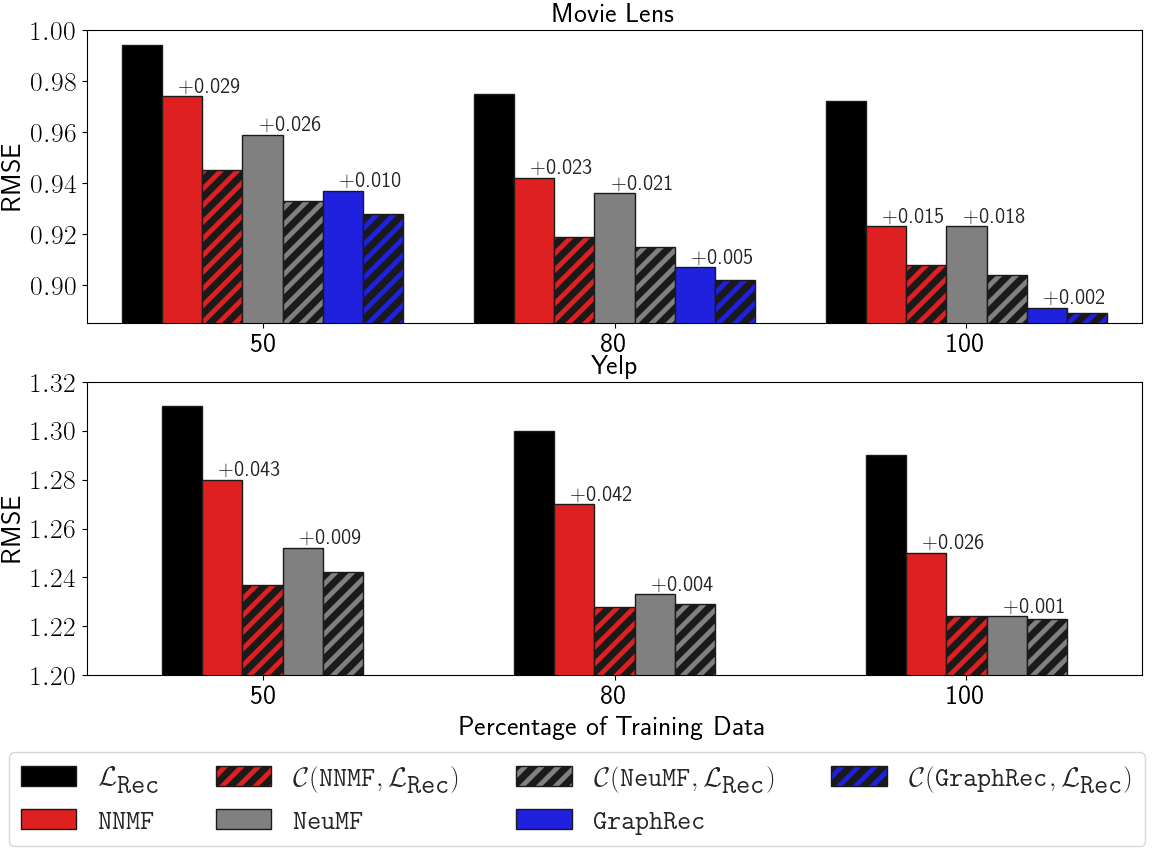}
    \caption{Results on a recommendation task on MovieLens (top) and Yelp (bottom).}
    \label{fig:reg_results}
\end{figure}

\textbf{Results}
Figure~\ref{fig:reg_results} shows the RMSE for different percentages of training data used (100\% is equivalent to the whole dataset). 
We measured performance in terms of Root Mean Squared Error (RMSE). The results are shown for $\theoryrec$ being trained independently of the neural components. Concordia can substantially improve the performance of the respective deep models in most cases.
For instance, when the full MovieLens-100k dataset is used the RMSE drops from 0.923 to 0.908 with regards to \nnmf;
with regards to \neumf, the RMSE drops from 
0.923 to 0.904 and even on \texttt{GraphRec}, a GNN designed specifically for recommendation systems, we improve from 0.891 to 0.889. We tested our hypothesis that the symbolic reasoning will increase even further when less data is available and found this to hold across models. In addition, we can see that except for \texttt{GraphRec} the Concordia models with 80\% data outperform their purely neural models with 100\%, and with 50\% data outperform the neural models with 80\% data. This shows that independent of the model, introducing symbolic rules further helps the neural models and reduces the need for data. The same behaviour can be observed on the Yelp dataset. \texttt{GraphRec} was not capable of performing training on this dataset due to its size. This shows that while GNNs can help in a similar fashion to symbolic models, as they represent relations well, they cannot be used in larger datasets where Concordia is still able to perform on other neural models.

\subsection{Collective activity detection}

CAD asks to identify the activity taking place by a group of actors in a collection of video frames \cite{collective-activity-2, collective-activity-3}. 
We used the dataset from \cite{choi2011learning}
and the train/test splits from \cite{collective-activity-3}. 
We denote by $\concordia(\mobilenet, \theorycad)$ and $\concordia$(\inception, $\theorycad$) the two different Concordia setups, where $\theorycad$ is taken from \cite{collective-psl}. 

\textbf{Results}
Table~\ref{tab:results_CAD} reports the average, best and worst accuracy over five different training/testing runs. 
Accuracy is defined as the percentage of the correctly predicted labels for group activity. 
In contrast to what was reported in \cite{arxiv2020} and despite that we used their code base, 
IARG was less effective than \mobilenet~ and \inception~ alone in most cases. 
Table~\ref{tab:results_CAD} shows that $\concordia(\mobilenet, \theorycad)$ improves over \mobilenet~ both in terms of best and  average accuracy. In particular, the average accuracy improves from
90.07\% to 90.73\%, while the best one improves from 
91.36\% to 93.19\%. 
With regards to \inception, the accuracy improvements brought by Concordia are even more significant. In particular, the average accuracy improves from 89.72\% to 92.75\%, the best accuracy from 91.83\% to 93.34\% and the worst accuracy from 86.84\% to 92.31\%. Notice that these results are state-of-the-art in CAD outperforming the ones from \cite{CVPR19,arxiv2020}. 
For completeness, Table~\ref{tab:results_CAD} also copies the results obtained when using $\theorycad$ in a stratified fashion over action context descriptors (ACD) \cite{ACD} for solving the task (ACD + $\theorycad$). We also ran the models on different dataset sizes by taking 50\%, 80\% and 100\% of the training data to test our hypothesis that the fewer data we have the larger the impact. Similarly to the recommendation task, we found that to hold across models, where on \mobilenet~ the addition of symbolic knowledge increase from 0.67\% improvement in accuracy to 1.27\%, while on the \inception~ model the improvement was from 3.03\% for 100\% data to 4.14\% improvement on 50\% of the data.

\begin{table}[!ht]
\centering
\caption{Results on CAD. }
\begin{tabular*}{\columnwidth}{l@{\extracolsep{\fill}}lll@{}}
\toprule
Model                               & AA (\%) & BA (\%) & WA (\%) \\ \midrule
ACD+$\theorycad$                               & 86.00 & - & - \\ \hline
\texttt{MNet}                       & 90.07            & 91.36         & 89.61          \\
\IARG(\texttt{MNet})                 & 90.18            & 92.39         & 87.55          \\
$\concordia$(\texttt{MNet}, $\theorycad$)    & 90.73            & 93.19        & {89.54}         \\ \hline
\texttt{Incept}                     & 89.72            & 91.83         & 86.84         \\ 
\IARG(\texttt{Incept})               & 88.88            & 91.67         & 85.33          \\
$\concordia$(\texttt{Incept}, $\theorycad$)    & \textbf{92.75}            & \textbf{93.34}         & \textbf{92.31}          \\ \bottomrule
\end{tabular*}
\label{tab:results_CAD}
\end{table}

\begin{table}[!ht]
\centering
\caption{Results on CAD over different dataset sizes. }
\begin{tabular*}{\columnwidth}{l@{\extracolsep{\fill}}lll@{}}
\toprule
Model                               & 50 \% & 80 \% & 90\% \\ \midrule
\texttt{MNet}                       & 68.70            & 77.1         & 90.07          \\
$\concordia$(\texttt{MNet}, $\theorycad$)    & 69.97            & 78.29        & 90.73         \\
Difference & 1.27 & 1.19 & 0.67\\ \hline
\texttt{Incept}                     & 71.10            & 84.55         & 89.72         \\ 
$\concordia$(\texttt{Incept}, $\theorycad$)    & \textbf{75.24}            & \textbf{88.59}         & \textbf{92.75}          \\
Difference & 4.14 & 4.04 & 3.03\\ \bottomrule
\end{tabular*}
\label{tab:results_CAD_data}
\end{table}
\subsection{Entity linking}

\begin{table}[tb]
\centering
\caption{{Results on entity linking}.}
\begin{tabular*}{\columnwidth}{l@{\extracolsep{\fill}}llll@{}}
\toprule
Model                                                                   & F$_1$ & Pr & Rec                & Acc (\%) \\ \midrule
$\theoryel$ (u) & 0.77& 0.80& 0.75& 74.7\\
\dpl(\bilstm) (u)                                                             & 0.76  & 0.68         & \textbf{0.86}               & 70.0        \\
$\concordia(\bilstm,\theoryel)$ (u)                           & \textbf{0.78}            & \textbf{0.84}         & 0.72     & \textbf{{75.8}}      \\ \hline
$\theoryel$ (sp) &0.76 & 0.85 & 0.69 & 75.2 \\
\bilstm~ (sp)                                                    & 0.74 &     0.58     &        \textbf{1.00}        & 58.5       \\ 
$\concordia(\bilstm,\theoryel)$ (sm)                        & 0.82            & 0.91         & 0.74     & \textbf{80.1}     \\
\bert~ (sp)                                                    & 0.88 &     \textbf{0.99}     &        0.78        & 88.5       \\ 
$\concordia(\bert,\theoryel)$ (sm)                        & \textbf{0.91}            & \textbf{0.99}         & 0.81     & \textbf{91.4}     \\
\bottomrule
\end{tabular*}
\label{tab:el_un}
\end{table}
To compare Concordia to $\dpl$ \cite{DBLP:conf/emnlp/WangP18}, we used the PubMedParsed dataset from \cite{moen2013distributional} and its extension from \cite{DBLP:conf/emnlp/WangP18}.
{We initially used the setup $\concordia(\bilstm,\theoryel)$, where 
$\theoryel$ is taken from \cite{DBLP:conf/emnlp/WangP18} and is encoded in PSL to have a direct comparison using the neural model used by the $\dpl$ authors. Then, we also used the setup $\concordia(\bert,\theoryel)$ to compare to more modern NLP models and understand how Concordia performs compared to large pre-trained neural models and to test whether Concordia only improves on simple neural models or also on large complex neural models.
Firstly, as in \cite{DBLP:conf/emnlp/WangP18}, we trained Concordia in an unsupervised fashion using only unlabelled data. Secondly, we used half of the labelled data to train Concordia in a semi-supervised fashion: in each epoch, we firstly trained the \bilstm~, \bert~ and $\theoryel$ in a supervised fashion (using the labelled data) and then trained \bilstm~ and \bert~ in an unsupervised fashion (using the unlabelled data). 
In the unsupervised case, all rule weights were set to 1. In the semi-supervised one, the rules were trained on half of the labelled data in advance.} In addition, we also considered \bilstm~ and \bert~ alone trained in a supervised fashion on the labelled data.
We also considered $\theoryel$ alone in two settings. 
In the unsupervised setting, $\theoryel$ performs predictions using 
rules which are all weighted 1. In the supervised case,   
$\theoryel$ is trained alone (i.e., independently of Concordia and the neural component)
using the labelled data. The final baseline is \bilstm~ regularized via DPL in an unsupervised fashion, denoted by DPL(\bilstm). All models were trained across five different data folds.

\textbf{Results}
Table~\ref{tab:el_un} reports the results of our experiments: 
(u) stands for unsupervised, (sm) for semi-supervised and (sp) for supervised learning.
For DPL, we copied the results from \cite{DBLP:conf/emnlp/WangP18}. 

DPL uses priors in their factors that were set based on information extracted from the labelled data-- that contradicts the statement in their paper that the training was unsupervised. 
In contrast, we do not provide these priors to Concordia 
setting the weights of the rules in $\theoryel$ to 1. 
Still, even with rules with untrained weights, Concordia outperforms DPL by 5.8\% in accuracy and by 0.02 in F$_1$. Concordia in an entirely unsupervised case (Table \ref{tab:el_un} row 3) outperforms \bilstm~ (Table \ref{tab:el_un} row 5) trained in a supervised fashion on the small labelled data set by 17.3\%. Our semi-supervised experiment outperforms the supervised \bilstm~ by 21.6\% and even \bert~ by 2.9\%, while the F$_1$ score improves by 0.08 and 0.03 respectively, proving again our hypothesis that Concordia can help significantly in settings with small amount of labelled data.

{DPL uses Markov Networks (and not MLNs) and takes, according to the authors, 2.5h to train in a cluster with 25 CPU cores and 1 GPU. 
Due to approximation techniques of probabilistic logical solvers, which have worst case polynomial time complexity \cite{psl-long, MLN}, Concordia is much more time-efficient taking under 6min to train on 1 GPU and 4 CPU cores.}

\section{Discussion and Conclusions}\label{sec:discussion}

\textbf{Theory learning.}
Concordia assumes that the formulas of the theory are given. 
Our assumption is along the lines of the assumptions made by several other neurosymbolic techniques such as \cite{deepproblog,neurasp,tsamoura2020neuralsymbolic}, as well as T-S and DPL. 
Despite that this assumption might be considered as restrictive, one could argue that purely neural teacher-student architectures \cite{distillation1,distillation2}, as well as deep learning techniques in general, also assume prior knowledge. In the former case, the prior knowledge is encoded into the teacher in subsymbolic form, while in the latter case, prior knowledge is encoded into the deep architecture. 
One option to circumvent this shortcoming if no rules are available is to use structure learning approaches \cite{feldstein2023principled, kok2010learning, boostr}, these are frameworks that aim at learning logical formulae from data and then pass these formulae into the logical model. However, weight learning can also serve as the means to learn the structure of the rules as previously shown in 
\cite{pLogicNet} in the context of learning knowledge graph embeddings: 
we can simply inject many arbitrary rules into the model (e.g., in the worst-case all possible rules of a specific form\footnote{This is similar to inductive logic programming \cite{DBLP:journals/jair/EvansG18}, where users specify patterns of rules to learn.}) and then at training time, the weights of the useless rules will drop to 0. In fact, this phenomenon was observed in our entity linking scenario. 
There, most rules turned out to be useless and their weights dropped to 0.

\textbf{Expressiveness.} We focus on LGMs as they offer a good balance between expressivity and complexity. Probabilistic frameworks such as ProbLog have been criticized for their high inference overhead \cite{sota1}. This overhead has been the reason for which prior neurosymbolic techniques such as \cite{DBLP:conf/eccv/ZhuFF14} have resorted to LGMs.     
Notice that LGMs support (ground) recursive rules as they are based on first-order-logic.  

\textbf{Beyond LGMs.}
We presented an instantiation of Concordia based on LGMs (see Section~\ref{section:preliminaries}). 
However,   
Concordia does not make any assumptions on the logical theory, 
as long as it abides by the generic interface presented here, see Section~\ref{section:core}.
Beyond lifted graphical models, Concordia, can also support languages like ProbLog.
Similarly to MLNs and PSL, ProbLog treats the ground atoms in the theory as RVs. 
Secondly, ProbLog supports both MAP (or MPE) inference \cite{fierens:tplp15} 
and supervised parameter learning \cite{learning-problog} 
and hence can implement Concordia's interfaces
${\problogical(Y  | \lveco, \lparams)}$ and 
${\update_\logical(\vec{x}, \vec{y}, \tau, \lparams_t) \rightarrow \lparams_{t+1}}$. 
We leave the integration of ProbLog as part of future work. 

\textbf{Conclusions.} 
We presented Concordia, a parallel neurosymbolic framework that is based on the formal semantics of probabilistic logical theories. 
Concordia can significantly improve the accuracy of deep models by injecting into them knowledge represented by probabilistic logic theories, leading to state-of-the-art results in a variety of tasks. Future work includes applying Concordia to other tasks, learning the structure of the rules at training time, and training the logical component in unsupervised settings.


\bibliography{concordia}

\begin{thebibliography}{54}
\providecommand{\natexlab}[1]{#1}
\providecommand{\url}[1]{\texttt{#1}}
\expandafter\ifx\csname urlstyle\endcsname\relax
  \providecommand{\doi}[1]{doi: #1}\else
  \providecommand{\doi}{doi: \begingroup \urlstyle{rm}\Url}\fi

\bibitem[Aditya et~al.(2019)Aditya, Yang, and Baral]{sota1}
Aditya, S., Yang, Y., and Baral, C.
\newblock Integrating knowledge and reasoning in image understanding.
\newblock In \emph{{IJCAI}}, pp.\  6252--6259, 2019.

\bibitem[Bach et~al.(2017)Bach, Broecheler, Huang, and Getoor]{psl-long}
Bach, S.~H., Broecheler, M., Huang, B., and Getoor, L.
\newblock Hinge-loss markov random fields and probabilistic soft logic.
\newblock \emph{Journal of Machine Learning Research}, 18:\penalty0
  109:1--109:67, 2017.

\bibitem[Choi et~al.(2011)Choi, Shahid, and Savarese]{choi2011learning}
Choi, W., Shahid, K., and Savarese, S.
\newblock Learning context for collective activity recognition.
\newblock In \emph{CVPR}, pp.\  3273--3280. IEEE, 2011.

\bibitem[Cohen et~al.(2020)Cohen, Yang, and
  Mazaitis]{DBLP:journals/jair/CohenYM20}
Cohen, W.~W., Yang, F., and Mazaitis, K.
\newblock Tensorlog: {A} probabilistic database implemented using deep-learning
  infrastructure.
\newblock \emph{J. Artif. Intell. Res.}, 67:\penalty0 285--325, 2020.
\newblock \doi{10.1613/jair.1.11944}.

\bibitem[Dai et~al.(2019)Dai, Xu, Yu, and Zhou]{abl}
Dai, W.-Z., Xu, Q., Yu, Y., and Zhou, Z.-H.
\newblock {Bridging Machine Learning and Logical Reasoning by Abductive
  Learning}.
\newblock In \emph{NeurIPS}, pp.\  2815--2826, 2019.

\bibitem[Dao et~al.(2021)Dao, Kamath, Syrgkanis, and Mackey]{distillation2}
Dao, T., Kamath, G.~M., Syrgkanis, V., and Mackey, L.
\newblock Knowledge distillation as semiparametric inference.
\newblock In \emph{ICLR}, 2021.

\bibitem[d'Avila Garcez et~al.(2002)d'Avila Garcez, Broda, and
  Gabbay]{DBLP:books/daglib/0007534}
d'Avila Garcez, A.~S., Broda, K., and Gabbay, D.~M.
\newblock \emph{Neural-symbolic learning systems: foundations and
  applications}.
\newblock Perspectives in neural computing. Springer, 2002.

\bibitem[De~Raedt \& Kimmig(2015)De~Raedt and Kimmig]{DeRaedt2015}
De~Raedt, L. and Kimmig, A.
\newblock Probabilistic (logic) programming concepts.
\newblock \emph{Machine Learning}, 100\penalty0 (1):\penalty0 5--47, 2015.

\bibitem[Devlin et~al.(2018)Devlin, Chang, Lee, and Toutanova]{bert}
Devlin, J., Chang, M.-W., Lee, K., and Toutanova, K.
\newblock Bert: Pre-training of deep bidirectional transformers for language
  understanding.
\newblock \emph{arXiv preprint arXiv:1810.04805}, 2018.

\bibitem[Donadello et~al.(2017)Donadello, Serafini, and d'Avila Garcez]{LTN}
Donadello, I., Serafini, L., and d'Avila Garcez, A.~S.
\newblock Logic tensor networks for semantic image interpretation.
\newblock In \emph{IJCAI}, pp.\  1596--1602, 2017.

\bibitem[Dziugaite \& Roy(2015)Dziugaite and Roy]{dziugaite2015neural}
Dziugaite, G.~K. and Roy, D.~M.
\newblock Neural network matrix factorization.
\newblock \emph{arXiv preprint arXiv:1511.06443}, 2015.

\bibitem[Evans \& Grefenstette(2018)Evans and
  Grefenstette]{DBLP:journals/jair/EvansG18}
Evans, R. and Grefenstette, E.
\newblock Learning explanatory rules from noisy data.
\newblock \emph{J. Artif. Intell. Res.}, 61:\penalty0 1--64, 2018.
\newblock \doi{10.1613/jair.5714}.

\bibitem[Feldstein et~al.(2023)Feldstein, Phillips, and
  Tsamoura]{feldstein2023principled}
Feldstein, J., Phillips, D., and Tsamoura, E.
\newblock Principled and efficient motif finding for structure learning in
  lifted graphical models.
\newblock \emph{arXiv preprint arXiv:2302.04599}, 2023.

\bibitem[Fierens et~al.(2015)Fierens, {Van den Broeck}, Renkens, Shterionov,
  Gutmann, Thon, Janssens, and {De Raedt}]{fierens:tplp15}
Fierens, D., {Van den Broeck}, G., Renkens, J., Shterionov, D.~S., Gutmann, B.,
  Thon, I., Janssens, G., and {De Raedt}, L.
\newblock Inference and learning in probabilistic logic programs using weighted
  boolean formulas.
\newblock \emph{Theory and Practice of Logic Programming ({TPLP})}, 15\penalty0
  (3):\penalty0 358--401, 2015.

\bibitem[Fischer et~al.(2019)Fischer, Balunovic, Drachsler-Cohen, Gehr, Zhang,
  and Vechev]{dl2}
Fischer, M., Balunovic, M., Drachsler-Cohen, D., Gehr, T., Zhang, C., and
  Vechev, M.
\newblock Dl2: training and querying neural networks with logic.
\newblock In \emph{International Conference on Machine Learning}, pp.\
  1931--1941. PMLR, 2019.

\bibitem[Ganchev et~al.(2010)Ganchev, Gra\c{c}a, Gillenwater, and
  Taskar]{posterior}
Ganchev, K., Gra\c{c}a, J.~a., Gillenwater, J., and Taskar, B.
\newblock Posterior regularization for structured latent variable models.
\newblock \emph{Journal of Machine Learning Research}, 11:\penalty0
  2001–2049, 2010.

\bibitem[Gutmann et~al.(2008)Gutmann, Kimmig, Kersting, and
  De~Raedt]{learning-problog}
Gutmann, B., Kimmig, A., Kersting, K., and De~Raedt, L.
\newblock Parameter learning in probabilistic databases: A least squares
  approach.
\newblock In \emph{Machine Learning and Knowledge Discovery in Databases}, pp.\
   473--488, 2008.

\bibitem[Harper \& Konstan(2015)Harper and Konstan]{10.1145/2827872}
Harper, F.~M. and Konstan, J.~A.
\newblock The movielens datasets: History and context.
\newblock \emph{ACM Transactions on Interactive Intelligent Systems},
  5\penalty0 (4), 2015.

\bibitem[He et~al.(2017{\natexlab{a}})He, Gkioxari, Doll{\'{a}}r, and
  Girshick]{roialign}
He, K., Gkioxari, G., Doll{\'{a}}r, P., and Girshick, R.~B.
\newblock Mask {R-CNN}.
\newblock In \emph{ICCV}, pp.\  2980--2988, 2017{\natexlab{a}}.

\bibitem[He et~al.(2017{\natexlab{b}})He, Liao, Zhang, Nie, Hu, and
  Chua]{he2017neural}
He, X., Liao, L., Zhang, H., Nie, L., Hu, X., and Chua, T.-S.
\newblock Neural collaborative filtering.
\newblock In \emph{WWW}, pp.\  173--182, 2017{\natexlab{b}}.

\bibitem[Hinton et~al.(2015)Hinton, Vinyals, and Dean]{distillation3}
Hinton, G.~E., Vinyals, O., and Dean, J.
\newblock Distilling the knowledge in a neural network.
\newblock \emph{CoRR}, abs/1503.02531, 2015.

\bibitem[Hu et~al.(2016{\natexlab{a}})Hu, Ma, Liu, Hovy, and
  Xing]{hu-etal-2016-harnessing}
Hu, Z., Ma, X., Liu, Z., Hovy, E., and Xing, E.
\newblock Harnessing deep neural networks with logic rules.
\newblock In \emph{ACL}, pp.\  2410--2420, 2016{\natexlab{a}}.

\bibitem[Hu et~al.(2016{\natexlab{b}})Hu, Yang, Salakhutdinov, and
  Xing]{hu-etal-2016-deep}
Hu, Z., Yang, Z., Salakhutdinov, R., and Xing, E.
\newblock Deep neural networks with massive learned knowledge.
\newblock In \emph{EMNLP}, pp.\  1670--1679, 2016{\natexlab{b}}.

\bibitem[Ibrahim \& Mori(2018)Ibrahim and Mori]{collective-activity-2}
Ibrahim, M.~S. and Mori, G.
\newblock Hierarchical relational networks for group activity recognition and
  retrieval.
\newblock In \emph{{ECCV}}, pp.\  742--758, 2018.

\bibitem[Jacobs et~al.(1991)Jacobs, Jordan, Nowlan, and
  Hinton]{jacobs1991adaptive}
Jacobs, R.~A., Jordan, M.~I., Nowlan, S.~J., and Hinton, G.~E.
\newblock Adaptive mixtures of local experts.
\newblock \emph{Neural computation}, 3\penalty0 (1):\penalty0 79--87, 1991.

\bibitem[Khot et~al.(2015)Khot, Natarajan, Kersting, and Shavlik]{boostr}
Khot, T., Natarajan, S., Kersting, K., and Shavlik, J.
\newblock Gradient-based boosting for statistical relational learning: the
  markov logic network and missing data cases.
\newblock \emph{Machine Learning}, 100\penalty0 (1):\penalty0 75--100, 2015.

\bibitem[Kok \& Domingos(2010)Kok and Domingos]{kok2010learning}
Kok, S. and Domingos, P.~M.
\newblock Learning markov logic networks using structural motifs.
\newblock In \emph{ICML}, volume~10, pp.\  551--558, 2010.

\bibitem[Koren et~al.(2009)Koren, Bell, and Volinsky]{koren2009matrix}
Koren, Y., Bell, R., and Volinsky, C.
\newblock Matrix factorization techniques for recommender systems.
\newblock \emph{Computer}, 42\penalty0 (8):\penalty0 30--37, 2009.

\bibitem[Kouki et~al.(2015)Kouki, Fakhraei, Foulds, Eirinaki, and
  Getoor]{kouki2015hyper}
Kouki, P., Fakhraei, S., Foulds, J., Eirinaki, M., and Getoor, L.
\newblock Hyper: A flexible and extensible probabilistic framework for hybrid
  recommender systems.
\newblock In \emph{RecSys}, pp.\  99--106, 2015.

\bibitem[Kuang \& Tie(2020)Kuang and Tie]{arxiv2020}
Kuang, Z. and Tie, X.
\newblock Video understanding based on human action and group activity
  recognition.
\newblock \emph{CoRR}, abs/2010.12968, 2020.

\bibitem[Lan et~al.(2012)Lan, Wang, Mori, and Robinovitch]{ACD}
Lan, T., Wang, Y., Mori, G., and Robinovitch, S.~N.
\newblock Retrieving actions in group contexts.
\newblock In \emph{ECCV Workshops}, pp.\  181--194, 2012.

\bibitem[Li \& Srikumar(2019)Li and Srikumar]{li2019augmenting}
Li, T. and Srikumar, V.
\newblock Augmenting neural networks with first-order logic.
\newblock \emph{arXiv preprint arXiv:1906.06298}, 2019.

\bibitem[{London} et~al.(2013){London}, {Khamis}, {Bach}, {Huang}, {Getoor},
  and {Davis}]{collective-psl}
{London}, B., {Khamis}, S., {Bach}, S.~H., {Huang}, B., {Getoor}, L., and
  {Davis}, L.
\newblock Collective activity detection using hinge-loss markov random fields.
\newblock In \emph{CVPR Workshops}, pp.\  566--571, 2013.

\bibitem[Manhaeve et~al.(2018)Manhaeve, Dumancic, Kimmig, Demeester, and
  De~Raedt]{deepproblog}
Manhaeve, R., Dumancic, S., Kimmig, A., Demeester, T., and De~Raedt, L.
\newblock Deepproblog: Neural probabilistic logic programming.
\newblock In \emph{NeurIPS}, pp.\  3749--3759, 2018.

\bibitem[Marra et~al.(2020)Marra, Diligenti, Giannini, Gori, and
  Maggini]{DBLP:conf/ecai/MarraDGGM20}
Marra, G., Diligenti, M., Giannini, F., Gori, M., and Maggini, M.
\newblock Relational neural machines.
\newblock In \emph{ECAI}, volume 325, pp.\  1340--1347, 2020.

\bibitem[Minervini et~al.(2018)Minervini, Bosnjak, Rockt{\"a}schel, and
  Riedel]{DBLP:journals/corr/abs-1807-08204}
Minervini, P., Bosnjak, M., Rockt{\"a}schel, T., and Riedel, S.
\newblock Towards neural theorem proving at scale.
\newblock \emph{arXiv preprint arXiv:1807.08204}, 2018.

\bibitem[Mobahi et~al.(2020)Mobahi, Farajtabar, and Bartlett]{distillation1}
Mobahi, H., Farajtabar, M., and Bartlett, P.~L.
\newblock Self-distillation amplifies regularization in hilbert space.
\newblock In \emph{NeurIPS}, 2020.

\bibitem[Moen \& Ananiadou(2013)Moen and Ananiadou]{moen2013distributional}
Moen, S. and Ananiadou, T. S.~S.
\newblock Distributional semantics resources for biomedical text processing.
\newblock \emph{Proceedings of LBM}, pp.\  39--44, 2013.

\bibitem[Ning et~al.(2015)Ning, Desrosiers, and Karypis]{Ning2015}
Ning, X., Desrosiers, C., and Karypis, G.
\newblock \emph{A Comprehensive Survey of Neighborhood-Based Recommendation
  Methods}, pp.\  37--76.
\newblock Springer, 2015.

\bibitem[Peng et~al.(2017)Peng, Poon, Quirk, Toutanova, and Yih]{peng2017cross}
Peng, N., Poon, H., Quirk, C., Toutanova, K., and Yih, W.-t.
\newblock Cross-sentence n-ary relation extraction with graph lstms.
\newblock \emph{Transactions of the Association for Computational Linguistics},
  5:\penalty0 101--115, 2017.

\bibitem[Qi et~al.(2018)Qi, Qin, Li, Wang, Luo, and
  Gool]{collective-activity-3}
Qi, M., Qin, J., Li, A., Wang, Y., Luo, J., and Gool, L.~V.
\newblock stagnet: An attentive semantic {RNN} for group activity recognition.
\newblock In \emph{{ECCV}}, pp.\  104--120, 2018.

\bibitem[Qu \& Tang(2019)Qu and Tang]{pLogicNet}
Qu, M. and Tang, J.
\newblock Probabilistic logic neural networks for reasoning.
\newblock In \emph{NeurIPS}, pp.\  7710--7720, 2019.

\bibitem[Rashed et~al.(2019)Rashed, Grabocka, and Schmidt-Thieme]{graphrec}
Rashed, A., Grabocka, J., and Schmidt-Thieme, L.
\newblock Attribute-aware non-linear co-embeddings of graph features.
\newblock In \emph{Proceedings of the 13th ACM conference on recommender
  systems}, pp.\  314--321, 2019.

\bibitem[Richardson \& Domingos(2006)Richardson and Domingos]{MLN}
Richardson, M. and Domingos, P.~M.
\newblock Markov logic networks.
\newblock \emph{Mach. Learn.}, 62\penalty0 (1-2):\penalty0 107--136, 2006.

\bibitem[Salakhutdinov \& Mnih(2008)Salakhutdinov and
  Mnih]{salakhutdinov2008bayesian}
Salakhutdinov, R. and Mnih, A.
\newblock Bayesian probabilistic matrix factorization using markov chain monte
  carlo.
\newblock In \emph{ICML}, pp.\  880--887, 2008.

\bibitem[Sanh et~al.(2019)Sanh, Debut, Chaumond, and Wolf]{distilbert}
Sanh, V., Debut, L., Chaumond, J., and Wolf, T.
\newblock Distilbert, a distilled version of bert: smaller, faster, cheaper and
  lighter.
\newblock \emph{arXiv preprint arXiv:1910.01108}, 2019.

\bibitem[Sikka et~al.(2020)Sikka, Silberfarb, Byrnes, Sur, Chow, Divakaran, and
  Rohwer]{DASL}
Sikka, K., Silberfarb, A., Byrnes, J., Sur, I., Chow, E., Divakaran, A., and
  Rohwer, R.
\newblock Deep adaptive semantic logic (dasl): Compiling declarative knowledge
  into deep neural networks.
\newblock \emph{arXiv preprint arXiv:2003.07344}, 2020.

\bibitem[Tsamoura et~al.(2021)Tsamoura, Hospedales, and
  Michael]{tsamoura2020neuralsymbolic}
Tsamoura, E., Hospedales, T., and Michael, L.
\newblock Neural-symbolic integration: A compositional perspective.
\newblock In \emph{AAAI}, 2021.

\bibitem[van Krieken et~al.(2020)van Krieken, Acar, and van
  Harmelen]{KR2020-92}
van Krieken, E., Acar, E., and van Harmelen, F.
\newblock {Analyzing Differentiable Fuzzy Implications}.
\newblock In \emph{KR}, pp.\  893--903, 2020.

\bibitem[Wang \& Poon(2018)Wang and Poon]{DBLP:conf/emnlp/WangP18}
Wang, H. and Poon, H.
\newblock Deep probabilistic logic: {A} unifying framework for indirect
  supervision.
\newblock In \emph{EMNLP}, pp.\  1891--1902, 2018.

\bibitem[Wu et~al.(2019)Wu, Wang, Wang, Guo, and Wu]{CVPR19}
Wu, J., Wang, L., Wang, L., Guo, J., and Wu, G.
\newblock Learning actor relation graphs for group activity recognition.
\newblock In \emph{{CVPR}}, pp.\  9964--9974, 2019.

\bibitem[Xie et~al.(2019)Xie, Xu, Meel, Kankanhalli, and
  Soh]{DBLP:conf/nips/XieXMKS19}
Xie, Y., Xu, Z., Meel, K.~S., Kankanhalli, M.~S., and Soh, H.
\newblock Embedding symbolic knowledge into deep networks.
\newblock In \emph{NeurIPS}, pp.\  4235--4245, 2019.

\bibitem[Yang et~al.(2020)Yang, Ishay, and Lee]{neurasp}
Yang, Z., Ishay, A., and Lee, J.
\newblock {NeurASP}: Embracing neural networks into answer set programming.
\newblock In \emph{{IJCAI}}, pp.\  1755--1762, 2020.

\bibitem[Zhu et~al.(2014)Zhu, Fathi, and Fei{-}Fei]{DBLP:conf/eccv/ZhuFF14}
Zhu, Y., Fathi, A., and Fei{-}Fei, L.
\newblock Reasoning about object affordances in a knowledge base
  representation.
\newblock In \emph{ECCV}, volume 8690, pp.\  408--424, 2014.

\end{thebibliography}
\bibliographystyle{icml2023}

%%%%%%%%%%%%%%%%%%%%%%%%%%%%%%%%%%%%%%%%%%%%%%%%%%%%%%%%%%%%%%%%%%%%%%%%%%%%%%%
%%%%%%%%%%%%%%%%%%%%%%%%%%%%%%%%%%%%%%%%%%%%%%%%%%%%%%%%%%%%%%%%%%%%%%%%%%%%%%%
% APPENDIX
%%%%%%%%%%%%%%%%%%%%%%%%%%%%%%%%%%%%%%%%%%%%%%%%%%%%%%%%%%%%%%%%%%%%%%%%%%%%%%%
%%%%%%%%%%%%%%%%%%%%%%%%%%%%%%%%%%%%%%%%%%%%%%%%%%%%%%%%%%%%%%%%%%%%%%%%%%%%%%%
\clearpage
\appendix
\section{Parallel neurosymbolic architectures}

Let $\mathcal{X}$ and $\mathcal{Y}$ denote the input and target domain, respectively,   
$(\vec{x}, \vec{y}) \in \mathcal{X} \times \mathcal{Y}$ denote a training datum and $\mathcal{D}$ denote the set of training data.  
Both T-S and DPL assume that the logical theory includes constraints of the form ${\mathcal{X} \times \mathcal{Y} \rightarrow [0,1]}$ where each constraint is associated with some level of confidence ${\lambda \in [0,1]}$. These constraints are functions checking the validity of a specific training datum (returning zero for no validity and one for perfect validity) and can be captured using graphical models. 

\subsection{The T-S framework}
Assuming that the neural model with parameters $\nparams$ defines a conditional probability distribution $p_{\nparams}(\vec{Y}|\vec{X} )$, T-S builds a teacher probabilistic model $q(\vec{Y}|\vec{X})$ and uses this network to update the parameters of the neural model $\nparams$. The teacher network is found by solving the following optimization problem: 

\begin{align}
\begin{split}
    & \min \limits_{q, \vec{\xi} \ge 0} KL((q(\vec{Y}|\vec{X}) || (\vec{Y}|\vec{X})) + C\sum_{l} \xi_{l}\\
&\text{s.t. }\lambda_l(1-\mathbb{E}[Z_l] \le \xi_{l}, \forall l 
\end{split}
\end{align}
 
In the above formulation, $C$ denotes a balancing parameter, $\xi_{l}$ denotes a slack variable along the lines of \cite{posterior} and $\vec{\xi}$ is the vector over all the $\xi_{l}$ occurring in the optimization objective. Let us define a random variable $Z_l$ with domain the set of possible confidences $\{r_l(\vec{x}, \vec{y})\}_{(\vec{x}, \vec{y}) \in \mathcal{D}}$ when provided in the input with different items from the training set, where the probability of each ${r_l(\vec{x}, \vec{y})}$ is ${q(\vec{Y}|\vec{X})}$. Then $\mathbb{E}[Z_l]$ denotes the expectation of $Z_l$. In words, the teacher probability function is one that stays close to the neural predictions (that is what the KL term accounts for) and fits the rules (that is what the second term accounts for).        

The T-S formulation makes several assumptions. First, the shape of the constraints $r_l$ does not allow expressing generic first order logic theories. For instance, T-S cannot support the rule ${}$, as the rule references atoms that do not occur in the training data. Secondly, the optimization objective does not abide by the semantics of lifted graphical models. In particular, in the optimal solution, the $\xi_l$’s become 0 so that $\mathbb{E}[Z_l]$ becomes 1. However, in such a case, the confidence of each constraint $\lambda_l$ is not taken into account.         

\subsection{The DPL framework}

DPL assumes that the training data is of the form ${\{\vec{x}_i\}_{i=1}^N}$.
Let $\Phi$ encode the probability distribution defined by the logical theory. Let us denote the conditional probability distribution
defined by
the neural model with parameters $\nparams$ by $p_{\nparams}(\vec{Y}|\vec{X} )$.
DPL defines the joint probability distribution:
\begin{align}
\begin{split}
    &P^{DPL}(Y_1,\dots,Y_N | X_1,\dots,X_N) =\\&\Phi(X_1,\dots,X_N | Y_1,\dots,Y_N) \cdot \prod_i p_\theta(Y_i|X_i))
\end{split}
 \label{eq:dpl-model}      
\end{align}
Above, $Y_i$ denotes a random variable with domain the possible labels of ${\vec{x}_i}$. 

DPL uses the distribution from \eqref{eq:dpl-model} to train the components. In particular, at each training step, DPL first computes the marginal distributions ${q_i(Y_i) = P^{DPL}(Y_1| X_1,\dots,X_N)}$. The product of those distributions over all $i$’s, where ${1 \leq i \leq N}$, defines a new probability distribution $q(Y_1,\dots,Y_N)$. The weights of the logical theory are updated by firstly, computing the KL divergence between $q(Y_1,\dots,Y_N)$ and $\Phi(Y_1,\dots,Y_N | X_1,\dots,X_N)$ and, secondly, updating the weights of $\Phi$ so that the KL divergence is minimized. The weights of the deep network are amended in the same fashion. 
   
The above formulation exposes several restrictions. Firstly, similarly to T-S, DPL does not straightforwardly extend to more expressive probabilistic logic theories as the constraints can be expressed as factors in MRFs (see section~\ref{section:preliminaries}). The extension to lifted graphical models is not discussed. Hence, it does not allow expressing information in latent variables. 
Secondly, the assumption of independence between the logic and the deep component is not justified. 

\section{Details on the empirical analysis}\label{app:experiments}

Concordia has been developed in PyTorch 1.10.
The logic component of each task was implemented using the pslpython library\footnote{{https://pypi.org/project/pslpython/}}.

\textbf{Computational environment}
All experiments ran on a Linux machine with a NVidia  GeForce GTX 1080 Ti GPUs, 64 Intel(R) Xeon(R) Gold 6130 CPUs, and 256GB of RAM.

\subsection{CAD} 

\textbf{Dataset} We used the {Collective Activity Augmented Dataset} (CAAD) \cite{choi2011learning}.
The dataset includes 44 video sequences with five different group activities (crossing, waiting, queuing, walking and talking) and six different individual actions (N/A, crossing, waiting, queuing, walking and talking). The group activity of each frame is defined as the activity 
followed by most of the actors in the frame. 
Each input datum $\vec{x}$ includes a video frame and a set of bounding boxes within the given frame. Each label datum $\vec{y}$ defines the action within each bounding box, as well as the activity in the frame. 
We used the train and test splits proposed in \cite{collective-activity-3}, namely choosing 2/3 of the video sequences for training and the rest for testing. 

\subsubsection{Logical component}\label{section:experiments:cad:theory}
Theory $\theorycad$ includes the rules: 
    \begin{align}
        \begin{split}
            &\lambda_1: \textsc{frame}(B, F) \wedge \textsc{flabel}(F, A) \rightarrow \textsc{doing}(B,A)\\
            &\lambda_2: \textsc{doing}(B_1,A) \wedge \textsc{close}(B_1, B_2) \rightarrow \textsc{doing}(B_2,A)\\
           &\lambda_3: \textsc{sequence}(B_1,B_2) \wedge \textsc{close}(B_1, B_2)\\& \rightarrow \textsc{same}(B_1,B_2)\\ &\lambda_4:\textsc{doing}(B_1,A) \wedge\textsc{same}(B_1, B_2)\rightarrow \textsc{doing}(B_2,A) \\
            & \lambda_5: \textsc{dnn}(B,A) \rightarrow \textsc{doing}(B,A) \nonumber
        \end{split} \nonumber
    \end{align}
    Atom $\textsc{close}(B_1,B_2)$ denotes that bounding boxes 
    $B_1$ and $B_2$ are close to each other.
    Atom $\textsc{doing}(B,A)$ denotes that 
    the actor within bounding box $B$ is doing action $A$.  
    Atom $\textsc{frame}(B, F)$ denotes that bounding box $B$ belongs to frame $F$, atom $\textsc{flabel}(F, A)$ denotes that the group activity of frame $F$ is $A$, atom $\textsc{sequence}(B_1, B_2)$ denotes that two bounding boxes are from two frames in a direct sequence of each other, and atom $\textsc{same}(B_1, B_2)$ denotes whether the actor within bounding boxes $B_1$ and $B_2$ is the same.
    The first rule states that the activity of an actor is the same with the activity of the frame.
    The second rule states that two actors that are close to each other perform the same activity.
    The third rule states that if two bounding boxes are from  a direct sequence of frames, and the two bounding boxes across the two frames are close to each other, then they describe the same actor.
    The fourth rule states that if the actor within two bounding boxes is the same, then it is likely that she is performing the same activity. 
    The last rule is as in Example~\ref{example:priors}.
    The above rules hold with some uncertainty which is captured by the $\lambda$ parameters. 
    
At training time, we provided instantiations of the predicates 
\textsc{frame}, \textsc{flabel}, \textsc{close}, \textsc{dnn}, \textsc{sequence} and 
\textsc{doing} using the training data. 
To instantiate the predicate \textsc{close} we used an RBF kernel to measure the closeness of the bounding boxes, which outputs a value in [0, 1]. 
To instantiate predicate \textsc{dnn}, we followed the steps in Section~\ref{section:core:concordia}.  
At testing time, we provided instantiations to all predicates but \textsc{doing}.  

Note that neither DPL nor T-S can implement this logic due to the latent variable $\textsc{same}$, as both expect all conclusions of each rule to be target atoms. In addition, they are not capable of handling this task due to its multi-task nature, as it predicts both labels for each bounding box in the frame, as well as the group activity of the frame.
\subsubsection{Neural component}
We considered the same architecture 
as the state-of-the-art \cite{CVPR19,arxiv2020}: 
$\backbone$. 
Component $\texttt{B}$ takes as input an image represented by a 2-dimensional vector of size 480x720 and creates a feature map for each input frame $f$ of size 57x87.
Given 
the feature map computed by $\texttt{B}$, $\texttt{RoIAlign}$ \cite{roialign} then outputs feature vectors for each bounding box within $f$ of size 1024. 
$\texttt{L}$ is a fully connected layer predicting the activity within each bounding box in $f$, as well as the group activity in $f$. 
To ensure a fair comparison with prior art, we used \mobilenet~ and \inception~ for $\texttt{B}$ ending up with two 
different instantiations of the neural architecture.
We will refer to those architectures as \mobilenet~ and \inception. 
The loss function used in this experiment is cross-entropy.

\textbf{Inputs/outputs} 
The inputs are a set of bounding boxes all given as coordinates within the frames, which in turn are passed into the neural model as 2-dimensional vectors of size 480x720. The outputs are soft-max distribution vectors over the possible activities for each bounding box and the group activity of the frame, which include crossing, waiting, queuing, walking, talking, dancing, jogging, and N/A.

\textbf{Training}
To train \mobilenet~ and \inception~  
we used a minibatch of size 1 and set the learning rate to 0.00001.

\subsubsection{Baselines}

We considered \mobilenet~ and \inception~ as baselines and trained them for 30 epochs each.
We also compared against the state-of-the-art architecture from \cite{arxiv2020}, which uses a Graph Convolutional Network on top of \mobilenet~ and \inception.  
The technique is referred to as IARG. 
We denote by {IARG}(\mobilenet) and {IARG}(\inception) the two different variants.
We used the implementation provided by the authors\footnote{{https://github.com/kuangzijian/Improved-Actor-Relation-Graph-based-Group-Activity-Recognition}.} and
used the hyper-parameters from \cite{CVPR19,arxiv2020}. 
{IARG}(\mobilenet) and {IARG}(\inception) were trained for 100 epochs as in \cite{arxiv2020}. 
As neither T-S nor DPL support the rules in $\theorycad$, 
we did not consider them as baselines. 

\subsection{Recommendation}
\textbf{Dataset} We considered Yelp (the academic version) and MovieLens-100k \cite{10.1145/2827872}. 
Yelp contains user ratings on local businesses, 
as well as information about business categories and friendships between users. 
The goal is to predict the ratings of the users on
businesses they haven’t rated yet. 
The dataset is updated on an annual basis by adding
different businesses. We used the version from 2021 and considered businesses only from Cambridge (US).
MovieLens-100k is a movie recommendation dataset
containing categorical information of movie genres and user occupations.

We used a 90\%/10\% training/test split for both Yelp and MovieLens-100k. For MovieLens-100k, we used the split 
provided by the publishers of the dataset. For Yelp, we created a random split. We used 30\% of the training data as the unobserved variables and 70\% of the training data as the observed variables of each dataset to learn the rules' weights.

\subsubsection{Logical component}\label{section:experiments:regression:theory}

PSL has already been successfully applied in recommendation \cite{kouki2015hyper}. We used the same rules in our experiments. Below, we describe the rules 
from $\theoryrec$ in detail (omitting the rules weights for clarity). 

The first two rules encode information about the item and user similarity: 
\begin{align*}
\begin{split}
     \textsc{SimilarUsers}_{sim}(U_1, U_2) \wedge &\textsc{Rating}(U_1, I)\\ &\rightarrow  \textsc{Rating}(U_2, I)\\
    \textsc{SimilarItems}_{sim}(I_1, I_2) \wedge &\textsc{Rating}(U,I_1)\\ &\rightarrow \textsc{Rating}(U,I_2)
\end{split}
\end{align*} 

The first rules states that if two users are similar, then they will give similar ratings to items. The second states that users will give similar ratings to similar items. These rules are also agnostic to the similarity measure used between users and between items. We adopted the same similarity metrics used by Kouki et al. \cite{kouki2015hyper}, namely Pearson, Cosine, latent cosine, latent euclidean-- the last two metrics are computed on latent vectors extracted using the Matrix Factorization approach on the rating matrix of all users and items \cite{Ning2015}. The first two similarity metrics are computed between rating vectors for each user and item. In addition, for item-based similarity, the adjusted cosine similarity is also used. To follow the authors, we also integrated content-based similarity. Content-based similarity included information, such as restaurant's category and user's occupation. The similarity measure used for content-based similarity is the Jaccard index. Predicates $\textsc{SimilarUsers}_{sim}$ and $\textsc{SimilarItems}_{sim}$ are binary predicates that take values of 1 if the first constant is one of the $k$-nearest neighbours of the second constant. In this experiment we used $k = 50$. Predicate $\textsc{Rating}$ takes values in the range [0, 1] and represents the normalized rating score user $U$ gave to an item $I$. 

The authors in \cite{kouki2015hyper} also included rules encouraging the predicted ratings to be close to an average user rating a user gives to items and an average item rating that is usually given to the item by all other users:
\begin{align*}
    \textsc{AverageUserRating}(U) \leftrightarrow \textsc{Rating}(U,I)
    \\
    \textsc{AverageItemRating}(U) \leftrightarrow \textsc{Rating}(U,I)
\end{align*} 

PSL can also leverage existing collaborative filtering methods to compute the ratings as defined in the rules above. To follow the original work, the theory $\theoryrec$ included the same most widely used collaborative filtering methods: matrix factorization (MF) \cite{koren2009matrix}, Bayesian probabilistic matrix factorization (BP) \cite{salakhutdinov2008bayesian}, and item-based collaborative filtering (IB):

\begin{align*}
    \textsc{Rating}_{MF}(U,I) \leftrightarrow \textsc{Rating}(U,I)\\
    \\
    \textsc{Rating}_{BP}(U,I) \leftrightarrow \textsc{Rating}(U,I) \\
    \\
    \textsc{Rating}_{\substack{IB}}(U,I) \leftrightarrow \textsc{Rating}(U,I) 
\end{align*}

Finally, $\theoryrec$ incorporated social network influences, like friends have similar tastes and thus give similar ratings:

\begin{align*}
\textsc{Friends}(U_1, U_2) \wedge \textsc{Rating}(U_1, I) \rightarrow  \textsc{Rating}(U_2, I)
\end{align*}

The predicates were instantiated as discussed in \cite{kouki2015hyper}.
Concretely, the \textsc{rating} predicates were instantiated directly from the data. For \textsc{AverageUserRating} and \textsc{AverageItemRating}, we computed the average as usual, while the similarity measurements were computed on the remaining data in the dataset, such as age, occupation and sex, using popular similarity measurements such as cosine, Pearson, latent cosine, and latent euclidean measurements.

Note that neither DPL nor T-S can implement these kind of rules due to the bi-directionality, of e.g. $\textsc{AverageUserRating}(U) \leftrightarrow \textsc{Rating}(U,I)$, as both expect all conclusions of the rules to be the target atoms.
\subsubsection{Neural component}

We used three state-of-the-art networks: \nnmf \cite{dziugaite2015neural}, \neumf \cite{he2017neural}, and \texttt{GraphRec} \cite{graphrec}. All networks learn embedding vectors for each user and item (business or movie).

\textbf{Inputs/outputs} 
The inputs to the networks are pairs of user/items.
The outputs are the item ratings. 

\textbf{Training}
The batch size used for \nnmf was 32, the learning rate 0.001, and the $L_2$ norm set to 0.01 for regularization, and for \neumf batch size was 16, learning rate 0.001, and the $L_2$ norm was 0.01.
The parameters for \texttt{GraphRec} were as follows: batch size = 1000, learning rate was set to 0.00003, the $L_2$ norm was set to 0.05 for the user features and 0.02 for the item features. In all three baseline models the optimizer was set to Adam with loss function set to RMSE. 

\subsubsection{Baselines}
We considered four different baselines: \nnmf, \neumf, \texttt{GraphRec} and the PSL theory $\theoryrec$ alone. The parameters were tuned according to the respective papers \cite{dziugaite2015neural, he2017neural, graphrec}.

\subsection{Entity linking}
\textbf{Dataset} We used the PubMedParsed data set originally put forth by \cite{moen2013distributional}, where we used the data generation and processing files from \cite{DBLP:conf/emnlp/WangP18} for a fair comparison. The goal in this data set is to predict which words in a text are mentions of protein names. The data contains 96k unlabelled data items, where each item is a sentence. In addition, for testing the performance of the unsupervised model, the authors in \cite{DBLP:conf/emnlp/WangP18}, provided a set of 12k labelled data items, where the mentions were labelled as protein or non-protein. 

\subsubsection{Logical component}

We used the theory from \cite{DBLP:conf/emnlp/WangP18} and in particular, the one using distant supervision (DS), data programming (DP) and joint inference (JI).

\subsubsection{Neural component}
We used a Bidirectional Long Short-Term Memory (\bilstm) recurrent neural network, originally proposed by \cite{peng2017cross}, and used the implementation from \cite{DBLP:conf/emnlp/WangP18} for a fair comparison. In addition, we compared the performance of Concordia to DistilBERT (\bert) \cite{distilbert} a distilled version of BERT \cite{bert} while retaining 97\% of its language understanding capabilies and being 60\% faster. The aim was to understand how Concordia performs with respect to large pre-trained models.

\textbf{Inputs/outputs} 
The inputs are mentions from sentences from the PubMedParsed dataset \cite{moen2013distributional}. The output is a prediction on whether the mention is a protein or a non-protein. To generate the training and testing inputs, we used the data generation scripts from \cite{DBLP:conf/emnlp/WangP18}.

\textbf{Training}
In the case of the \bilstm, the embedding layer is initialized with a word2vec embedding trained on PubMed abstracts and entire texts. The word embedding dimension was 200 as in \cite{DBLP:conf/emnlp/WangP18}. We used a learning rate of 0.001 and batch size 64. In the case of \bert, we used the word embedding provided by the model itself as it is a pretrained model. We used a learning rate of 0.00003 and batch size 16. The loss function used in all experiments was cross-entropy.

\subsubsection{Baselines}
We were not able to reproduce the results reported in \cite{DBLP:conf/emnlp/WangP18} due to the exponential approach proposed by the authors to compute their factor graphs. The authors ran their experiments on clusters. Therefore, the results on DPL reported in the main body are taken from their paper. 

\section{Connection from LGMs to PSL}
We demonstrate the notions of LGMs and PSL by example.

\begin{example}  

Consider the second rule $r$ from Section \ref{section:experiments:cad:theory}:
\begin{align}
\textsc{doing}(B_1,A) \wedge \textsc{close}(B_1, B_2) \rightarrow \textsc{doing}(B_2,A) \nonumber
\end{align}

Assuming that there are two bounding boxes in total, $\mathtt{b_1}$ and $\mathtt{b_2}$, and one activity, $\mathtt{crossing}$, rule $r$ can be instantiated in two different ways:  

\begin{align}
\overset{X_{1,1}}{\textsc{doing}(\mathtt{b_1, crossing})} \wedge \overset{X_{1,2}}{\textsc{close}(\mathtt{b_1, b_2})} \rightarrow \nonumber \\ \overset{X_{1,3}}{\textsc{doing}(\mathtt{b_2, crossing})} \label{example:inst1} \\
\overset{X_{1,3}}{\textsc{doing}(\mathtt{b_2, crossing})} \wedge \overset{X_{2,2}}{\textsc{close}(\mathtt{b_2, b_1})} \rightarrow \nonumber \\ \overset{X_{1,1}}{\textsc{doing}(\mathtt{b_1, crossing})} \label{example:inst2} 
\end{align}
Above each ground atom we show the RV associated with it, e.g., ground atom ${\textsc{doing}(\mathtt{b_1, crossing})}$ is associated with the RV $X_{1,1}$, while ground atom ${\textsc{close}(\mathtt{b_1, b_2})}$ is associated with RV $X_{1,2}$. The domain of all RVs is ${[0,1]}$. 

We denote by $f_1$ and $f_2$ the factors associated with the rule instantiations 
\eqref{example:inst1} and \eqref{example:inst2}, respectively.
Let ${\vec{X}_1=(X_{1,1}, X_{1,2}, X_{1,3})}$ and ${\vec{X}_2=(X_{1,3}, X_{2,2}, X_{1,1})}$.
In PSL, each factor $f_i$ is defined as follows
\begin{align}
    f_i(\vec{X}_i = \vec{x}_i) = e^{-\lambda \cdot (1 -r(\vec{x}_i))^p}, 
\end{align}
where $\vec{x}_i$ an instantiation of $\vec{X}_i$, for ${i=1,2}$, 
and $p$ is as defined in \eqref{eq:pslfactor}.
The par-factor $\phi$ associated with rule $r$ is essentially the set ${\{f_1,f_2\}}$. We use 
$\phi(\vec{X}_i = \vec{x}_i)$ to denote $f_i(\vec{X}_i = \vec{x}_i)$, for ${i=1,2}$.

Let ${\vec{X} = (X_{1,1}, X_{1,2}, X_{2,2}, X_{1,3})}$ denote the vector composed over all RVs associated with ground atoms in the Herbrand base of our theory. 
Probability ${P(\vec{X} = \vec{x})}$ induced by the par-factor graph $\{\phi\}$, where ${\vec{x} = (x_{1,1}, x_{1,2}, x_{2,2}, x_{1,3})}$ is an instantiation of $\vec{X}$, is defined as follows: 
\begin{align}\label{eq:parFactorProb}
    P({\vec{X}} = {\vec{x}}) = \nonumber \frac{1}{Z} \phi((x_{1,1}, x_{1,2}, x_{1,3})) \cdot \phi((x_{1,3}, x_{2,2}, x_{1,1}))
\end{align}
where $Z$ is the normalization constant.
\end{example}
\section{Details on inference in the logical component}\label{app:logical}
In the main body of this paper, we touched on the differences of inference in the logical component of Concordia. In this section, we are revisiting inference in classification tasks, by considering the CAD example. Then, we will elaborate on the differences in regression.

\subsection{Classification}
We elaborated in the main body of this paper on inference in classification tasks, where we discussed the difference between the logical component admitting Boolean interpretations vs interpretations in [0,1].

\begin{example}
Consider the task of collective activity detection, described in Section \ref{section:experiments:cad:theory}. In this task, the $\textsc{doing}$ atoms, are what we referred to as \textit{target atoms} in the main body, as the labels for our data are the activities of the different bounding boxes in a frame.

\textbf{Boolean interpretation} We mentioned in the case of Boolean interpretation of the atoms, where the atoms are mapped to \textit{true} or \textit{false}, we obtain a probability for each option of $c_j \in \mathcal{Y}$. That is, we get for a bounding box $\texttt{B}_i$ $P(\textsc{doing}(\texttt{B}_i, \texttt{dancing}) = \texttt{true}) = p_1$, $P(\textsc{doing}(\texttt{B}_i, \texttt{talking}) = \texttt{true}) = p_2$, ...
However, obviously, only one activity can be true at a time, and therefore, $p_1 + p_2 + \dots + p_n =1$. To enforce this, we need to add an additional rule to the rules mentioned in Section \ref{section:experiments:cad:theory}, enforcing the mutual exclusiveness
\begin{align*}
    \textsc{doing}(\texttt{B}_i, \texttt{dancing}) \veebar \textsc{doing}(\texttt{B}_i, \texttt{talking}) \veebar  \dots
\end{align*}

\textbf{Soft interpretation} When the logical interpretations accept values in $[0,1]$, we have in effect, a probability distribution for each target atom, i.e. one distribution for $\textsc{doing}(\texttt{B}_i, \texttt{dancing})$, $\textsc{doing}(\texttt{B}_i, \texttt{talking})$,  ... Then, as described in Section \ref{section:core:logical}, we choose the probability of each class, as the MAP over the distribution of the soft truth values. Finally, the values need to be normalized such that their sum is 1 to enforce the properties of a probability distribution. This is done through the following rule
\begin{align*}
    \textsc{doing}(\texttt{B}_i, \texttt{dancing}) + \textsc{doing}(\texttt{B}_i, \texttt{talking}) +  \dots = 1
\end{align*}

\begin{figure}[ht]
    \centering
    \includegraphics[width=0.75\linewidth]{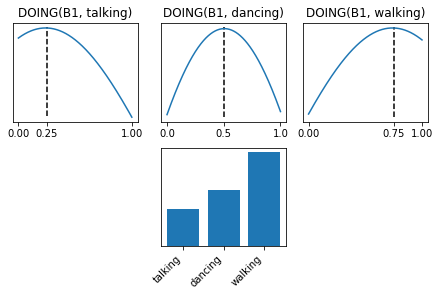}
    \caption{Illustrative example how a distribution over different labels can be obtained using PSL. Top row: each target atom has a distribution over their soft truth values. Bottom row: taking their respective MAPs and normalizing so that their sum is 1, produces a discrete distribution over the classes}
\label{fig:dist}
\end{figure}

Figure \ref{fig:dist} illustrates how a discrete distribution is obtained for the CAD example, where only three distinct actions are considered for illustrative purposes.
\end{example}

\subsection{Regression}

We support regression tasks via PSL. Recall that each target atom maps to a continuous RV $Y$ in ${[0,1]}$, i.e. the soft truth value (see Section~\ref{section:preliminaries}). 
The conditional probability ${\problogical(Y=y | \lveco, \lparams)}$ denotes the likelihood the target atom $Y$ takes soft truth value $y$.
We define $\problogical$ by marginalizing 
\begin{align*}
    P({\vec{X} = \vec{x}}) = \frac{1}{Z} \exp \left( \sum_{i=1}^{M} \sum_{j=1}^{M_{i}} \lambda_i f_{i,j}(\vec{X}_{i,j} = \vec{x}_{i,j}) \right),
\end{align*}
over all the remaining RVs. In case the domain of the regression task does not coincide with the soft truth domain {[0,1]} in PSL, the soft truth values are scaled to the target domain.  

Note that no additional constraint needs to be provided in regression tasks, as the distribution over the soft truth values already satisfies the axioms of a probability distribution.

\begin{example}
Let us consider the recommendation task, with the logical rules as described in Section \ref{section:experiments:regression:theory}. In this case, the target atom is $\textsc{rating}$, which takes values between 1 and 5. To achieve this, we scale the interval [1,5] onto the interval [0,1], such that the soft truth value of the $\textsc{rating}$ atoms equals the predicted rating.
\end{example}
%%%%%%%%%%%%%%%%%%%%%%%%%%%%%%%%%%%%%%%%%%%%%%%%%%%%%%%%%%%%%%%%%%%%%%%%%%%%%%%
%%%%%%%%%%%%%%%%%%%%%%%%%%%%%%%%%%%%%%%%%%%%%%%%%%%%%%%%%%%%%%%%%%%%%%%%%%%%%%%

\end{document}